# Domain Adaptation Framework for Turning Movement Count Estimation with Limited Data

**Xiaobo Ma, Ph.D.**
Pima Association of Governments
1 E Broadway Blvd., Ste. 401
Tucson, AZ 85701
Email: xiaoboma@arizona.edu

**Hyunsoo Noh, Ph.D.**
Pima Association of Governments
1 E Broadway Blvd., Ste. 401
Tucson, AZ 85701
Email: HNoh@pagregion.com

**Ryan Hatch**
Pima Association of Governments
1 E Broadway Blvd., Ste. 401
Tucson, AZ 85701
Email: RHatch@pagregion.com

**James Tokishi**
Pima Association of Governments
1 E Broadway Blvd., Ste. 401
Tucson, AZ 85701
Email: JTokishi@pagregion.com

**Zepu Wang**
Civil & Environmental Engineering
University of Washington
3760 E. Stevens Way NE
Seattle, WA 98195
Email: zepu@uw.edu

## ABSTRACT

Urban transportation networks are vital for the efficient movement of people and goods, necessitating effective traffic management and planning. An integral part of traffic management is understanding the turning movement counts (TMCs) at intersections, Accurate TMCs at intersections are crucial for traffic signal control, congestion mitigation, and road safety. In general, TMCs are obtained using physical sensors installed at intersections, but this approach can be cost-prohibitive and technically challenging, especially for cities with extensive road networks. Recent advancements in machine learning and data-driven approaches have offered promising alternatives for estimating TMCs. Traffic patterns can vary significantly across different intersections due to factors such as road geometry, traffic signal settings, and local driver behaviors. This domain discrepancy limits the generalizability and accuracy of machine learning models when applied to new or unseen intersections. In response to these limitations, this research proposes a novel framework leveraging domain adaptation (DA) to estimate TMCs at intersections by using traffic controller event-based data, road infrastructure data, and point-of-interest (POI) data. Evaluated on 30 intersections in Tucson, Arizona, the performance of the proposed DA framework was compared with state-of-the-art models and achieved the lowest values in terms of Mean Absolute Error and Root Mean Square Error.

**Keywords:** Turning Movement Counts, Domain Adaptation, Traffic Flow Estimation

# 1. INTRODUCTION

Urban transportation networks are the lifelines of modern cities, supporting the movement of people and goods. Efficient traffic management and planning are paramount for ensuring smooth and safe travel within these networks(Cottam et al., 2024; Ma, Karimpour, et al., 2023b). The accurate estimation of Turning Movement Counts (TMCs) at signalized intersections is a critical component in urban traffic management and planning. TMCs provide essential data for traffic signal control, congestion mitigation, and road safety enhancement. Traditional methods of collecting TMC data often involve manual counting, which can be costly, time-consuming, and susceptible to human error. As cities grow and traffic patterns become more complex, there is a pressing need for more efficient, reliable, and scalable methods to estimate TMCs.

Loop detectors, which are embedded in the pavement to detect the presence of vehicles, are commonly used for this purpose due to their reliability and relatively low cost (Ma et al., 2020). Ghods and Fu (2014) proposed a method to estimate TMCs at signalized intersections based on entry/exit traffic volumes collected from loop detectors and signal phase information (Ghods & Fu, 2014). Gholami and Tian (2016) used a network equilibrium approach to process loop detector data for estimating TMCs at shared lanes(Gholami & Tian, 2016). In another study, loop detector data combined with an adaptive neural fuzzy inference system and genetic programming were applied to estimate TMCs (Gholami & Tian, 2017) . A recent study employed data collected from conventional long-loop detectors for left-turn movement count estimation based on machine-learning models (Biswas et al., 2022). However, implementing these methods on a region-wide scale poses significant challenges. One of the primary obstacles is that the majority of intersections are not equipped with loop detectors. Without these detectors, accurately estimating TMCs at numerous intersections becomes an unfeasible endeavor(Wu et al., 2019).

The application of video cameras and computer vision technologies for estimating TMCs at intersections has gained substantial interest in recent years. Video cameras offer a versatile and non-intrusive means of capturing real-time traffic data, providing a comprehensive view of vehicular movements. Coupled with advancements in computer vision, these systems can automatically detect and track vehicles, classify them by type, and accurately count turning movements. Studies have demonstrated that computer vision algorithms, including convolutional neural networks (CNNs) and deep learning techniques, can achieve high levels of accuracy in traffic flow analysis. For instance, research by Shirazi and Morris (2016) highlighted the efficacy of using a vision-based vehicle tracking system to analyze video footage and estimate TMCs (Shirazi & Morris, 2016). Bélisle et al. (2017) introduced an innovative method for automatically counting vehicle turning movements using video tracking. It builds upon previous research that focused on optimizing parameters for extracting road user trajectories and automating the clustering of these trajectories (Bélisle et al., 2017). Adl et al. (2024) proposed an innovative framework for vehicle counting at intersections using a fisheye camera system. Their experiments on real-world datasets yielded promising results, underscoring the framework's potential for applications in intelligent traffic control and urban planning (Adl et al., 2024). Despite the advantages brought by video cameras and computer vision technologies, several challenges persist, such as the necessity for high-quality video footage, the development of robust algorithms capable of performing reliably under diverse weather conditions and lighting, and the significant computational resources required for real-time processing. Additionally, the high costs associated with advanced video and sensor technologies present significant financial barriers, preventing broad deployment despite their potential benefits. Thus, the widespread adoption of video cameras for TMC estimation is currently not feasible for many transportation agencies, necessitating the exploration of more cost-effective and scalable alternatives(Ma, Karimpour, et al., 2023a; Yang et al., 2024).

Recent advancements in machine learning and data-driven approaches have offered promising alternatives for estimating TMCs. Various studies have explored the use of supervised learning techniques, such as regression models and neural networks, to predict vehicle movements based on historical traffic data and real-time inputs(Ma, Karimpour, et al., 2023c; Ma, Noh, et al., 2024; Z. Zhang et al., 2024). For example, Ghanim and Shaaban (2019) utilized artificial neural networks (ANN) to analyze the relationship between approach volumes and the corresponding turning movements, leveraging large datasets to train their algorithms (Ghanim & Shaaban, 2019). Another study involves the use of support vector regression (SVR), random forest (RF), and ANN to improve estimation accuracy and robustness (Biswas et al., 2022). Xu et al. (2023) demonstrated that the well-calibrated ANN could effectively estimate TMCs by integrating traffic controller event-based data (Xu et al., 2023). Machine learning methods are scalable and can be deployed across multiple intersections, offering a cost-effective solution compared to traditional methods such as manual counting or loop detectors, as well as emerging technologies like video cameras and computer vision. Despite these advancements, conventional machine learning models often assume that the data distributions of training and testing datasets are identical, which is seldom the case in real-world scenarios. Traffic patterns can vary significantly across different intersections due to factors such as road geometry, traffic signal settings, and local driver behaviors. This domain discrepancy limits the generalizability and accuracy of traditional models when applied to new or unseen intersections(Ma, 2022; Ma, Cottam, et al., 2023; Z. Wang et al., 2025).

To reduce domain discrepancy and increase model generalizability, the concept of domain adaptation (DA) has gained significant traction in recent years. DA is an advanced machine learning technique that leverages knowledge from related tasks or domains to improve the performance of a model on a target task. By relaxing the assumption that the source and target data distributions must be identical, DA can adapt pre-trained models to new domains with limited new domain data, thus enhancing prediction accuracy and generalizability (Ma, Karimpour, et al., 2024). In the context of traffic management, DA can harness data from data-rich cities or intersections (source domain) to infer traffic patterns in cities or intersections with limited data (target domain). For example, Wang et al. (2022) proposed a Deep Attentive Adaptation Network, to transfer spatial-temporal knowledge across domains. The model maps source and target domain data to a common embedding space and uses domain adaptation with a discrepancy penalty to align domain distributions. The global attention mechanism captures complex spatial dependencies (S. Wang et al., 2022). In another study, Huang et al. (2023) presented TrafficTL, a cross-city traffic prediction framework that leverages big data from larger cities to improve prediction accuracy in data-scarce cities. By incorporating a periodicity-based transfer paradigm, it mitigates negative transfer caused by differences in data distributions and applies graph reconstruction techniques to address data deficiencies (Huang et al., 2023). In addition, Yao et al. (2023) introduced a traffic prediction model that is specifically designed for data-scarce road networks by transferring knowledge from data-rich networks. By combining spatial-temporal graph convolutional networks with adversarial domain adaptation, the model learns domain-invariant features to enable effective knowledge transfer (Yao et al., 2023). Moreover, Mo and Gong (2023) proposed a Cross-city Multi-Granular Adaptive Transfer Learning method by using limited target city data. The model employs meta-learning to initialize training on multiple source cities and extracts multi-granular regional features. An Adaptive Transfer module with Spatial-Attention and Multi-head Attention mechanisms is used to select and transfer the most relevant features (Mo & Gong, 2023). More recently, Li et al. (2024) introduced a macroscopic fundamental diagram (MFD) guided transfer learning method that identifies and transfers domain-invariant traffic flow patterns to address challenges like data insufficiency and dataset shift. Using an MFD similarity measure, the method determines transferable patterns and incorporates this measure into domain adversarial pre-training for improved adaptability (Li et al., 2024). Although the above-mentioned studies demonstrated satisfactory performance in traffic prediction and estimation, they often impose stringent requirements on both the volume and quality of data, as well as on computational resources. Furthermore, these methods rely on the availability of at least some labeled data from the target domain to achieve effective knowledge transfer and adaptation. Consequently, their applicability becomes significantly constrained in scenarios where labeled data in the target domain is completely unavailable, limiting their practicality in real-world settings where data labeling is expensive, time-consuming, or infeasible.

To overcome the above-mentioned limitations, this research proposes a domain adaptation (DA) framework for estimating Traffic Movement Counts (TMCs) when labeled data in the source domain is limited, and labeled data in the target domain is completely unavailable. The process begins with feature extraction to identify relevant traffic parameters and transform raw data into structured, usable information. Next, a feature selection technique, referred to as Least Absolute Shrinkage and Selection Operator (Lasso Regression), is applied to identify significant features, reducing data dimensionality and improving model performance. To ensure effective knowledge transfer, the Information-theoretic metric learning (ITML) algorithm matches target domain data to similar source domain data based on the selected features. To address the data scarcity problem, Gaussian Mixture Models (GMM) are employed to generate synthetic data that mimic the statistical properties of the matched labeled source domain data, helping to bridge domain gaps. The augmented source domain data is then substituted for the unavailable labeled target domain data, which is required for DA model training. This substitution process effectively creates a synthetic dataset for the target domain, enabling the use of existing data to infer traffic patterns in the absence of physical sensors. Finally, the Gradient Boosting with Balanced Weighting (GBBW) model is trained, assigning higher weights to labeled data instances similar to the augmented data, thereby prioritizing relevant information and enhancing regressor accuracy. The framework ultimately predicts TMCs, providing valuable insights for traffic management, including signal optimization, congestion reduction, and strategic planning.

The proposed framework accurately estimates TMCs at intersections by using traffic controller event-based data, road infrastructure data, and point-of-interest (POI) data. In addition, the proposed framework develops scene-specific models by employing a DA model. Because the DA model relaxes the assumption that the underlying data distributions of the source and target domains must be the same, the proposed framework can guarantee high-accuracy estimation of TMCs for intersections with different traffic patterns, distributions, and characteristics. The performance of the proposed framework is evaluated using 30 intersections in Tucson, Arizona.

This research makes several key contributions to the field of traffic management and TMC estimation:

- **Introduction of a novel DA framework for TMC estimation:** To the best of our knowledge, this study pioneers the application of DA in the context of TMC estimation, demonstrating its potential to improve accuracy and efficiency. This study proposes GBBW as an extension of Gradient Boosting by integrating the Balanced Weighting technique for domain adaptation. By emphasizing samples from the source domain that align closely with the target distribution, it effectively handles domain shifts while leveraging Gradient Boosting's strong predictive performance. This method enhances model generalization to target tasks and improves accuracy in scenarios with distribution differences between source and target domains.
- **The DA framework is rigorously compared to state-of-the-art models to highlight its benefits:** The proposed DA framework is rigorously compared with various state-of-the-art machine learning regression models and DA models on a real-world, large-scale road network for model validation. This comparison highlights its advantages and addresses potential challenges when labeled data in the source domain is limited and completely unavailable in the target domain.
- **This research proposes a scene-specific, scalable, cost-effective solution using traffic controller event data to continuously adapt to dynamic urban traffic conditions**: By tailoring models to specific intersections, the proposed DA framework provides precise, scene-specific TMC estimates while offering a scalable solution that can adapt to evolving traffic conditions. Urban traffic systems are highly dynamic, and the framework's ability to continuously update and adapt to new data ensures effective traffic monitoring and management. This research primarily leverages traffic controller event-based data, which is both cost-effective and widely available, as most local transportation agencies have already configured traffic detectors for actuated signal control, providing region-wide coverage without requiring additional infrastructure.
- **The proposed DA framework promotes data efficiency by enabling models to learn from limited data using information from other intersections:** Collecting extensive TMC data or implementing physical sensors at every intersection is both impractical and cost-prohibitive. The proposed DA framework addresses these challenges by enabling models to learn from limited data, leveraging related information from other intersections. This reduces the dependency on large datasets while maximizing the utility of existing data, offering a cost-effective alternative to extensive data collection and additional infrastructure.

In conclusion, the proposed DA framework represents a significant advancement in the estimation of TMCs. By addressing the limitations of conventional models and leveraging the strengths of DA, this research aims to provide a robust, efficient, and scalable solution for modern urban traffic management.

The remainder of this paper is organized as follows: preliminaries and problem definition are described in the second section. The third section introduces the proposed framework and methodologies. In the fourth section, a case study is presented to examine the transferability of the proposed framework. Lastly, section five provides the conclusion and future work.

## 2. PROBLEM DEFINITION AND PRELIMINARIES

The notations, traffic variable definitions, and data encoding are presented first. Subsequently, the problem addressed in this study is formulated.

### 2.1. Notations

Table 1 lists all the variables and their definitions that were used for model development.

**Table 1 Notations and Variables**

| Categories | Variables | Definition |
|---|---|---|
| Event information for through movement | $o_{TM}^{a,i,t}$ | Detector occupancy time for through movement of approach $a$ of intersection $i$ at time interval $t$ |
| | $d_{TM}^{a,i,t}$ | Detector trigger counts for through movement of approach $a$ of intersection $i$ at time interval $t$ |
| | $g_{TM}^{a,i,t}$ | Green time duration for through movement of approach $a$ of intersection $i$ at time interval $t$ |
| | $c_{TM}^{a,i,t}$ | Cycle counts for through movement of approach $a$ of intersection $i$ at time interval $t$ |
| | $m_{TM}^{a,i,t}$ | Average of time differences between each pair of consecutive detections for through movement of approach $a$ of intersection $i$ at time interval $t$ |
| | $s_{TM}^{a,i,t}$ | Standard deviation of time differences between each pair of consecutive detections for through movement of approach $a$ of intersection $i$ at the time interval $t$ |
| Event information for left-turn movement | $o_{LM}^{a,i,t}$ | Detector occupancy time for left-turn movement of approach $a$ of intersection $i$ at time interval $t$ |
| | $d_{LM}^{a,i,t}$ | Detector trigger counts for left-turn movement of approach $a$ of intersection $i$ at time interval $t$ |
| | $g_{LM}^{a,i,t}$ | Green time duration for left-turn movement of approach $a$ of intersection $i$ at time interval $t$ |
| | $c_{LM}^{a,i,t}$ | Cycle counts for left-turn movement of approach $a$ of intersection $i$ at time interval $t$ |
| | $m_{LM}^{a,i,t}$ | Average of time differences between each pair of consecutive detections for left-turn movement of approach $a$ of intersection $i$ at time interval $t$ |
| | $s_{LM}^{a,i,t}$ | Standard deviation of time differences between each pair of consecutive detections for left-turn movement of approach $a$ of intersection $i$ at the time interval $t$ |
| | $p_{LM}^{a,i,t}$ | Permissive green time for left-turn movement of approach $a$ of intersection $i$ at time interval $t$ |
| Intersection layout information | $l_{SL}^{a,i}$ | Number of shared left turn lanes of approach $a$ of intersection $i$ |
| | $l_{EL}^{a,i}$ | Number of exclusive left turn lanes of approach $a$ of intersection $i$ |
| | $l_{TL}^{a,i}$ | Number of through lanes of approach $a$ of intersection $i$ |
| | $l_{ER}^{a,i}$ | Number of exclusive right turn lanes of approach $a$ of intersection $i$ |
| | $l_{SR}^{a,i}$ | Number of shared right turn lanes of approach $a$ of intersection $i$ |
| POI information | $e_{POIE}^{i}$ | Number of employees of all POI within 400 m of intersection $i$ |
| | $e_{POIC}^{i}$ | POI categories count within 400 m of intersection $i$ |
| Turning movement counts | $v_{LM}^{a,i,t}$ | Traffic counts for left-turn movement of approach $a$ of intersection $i$ at the time interval $t$ |
| | $v_{TM}^{a,i,t}$ | Traffic counts for through movement of approach $a$ of intersection $i$ at the time interval $t$ |
| | $v_{RM}^{a,i,t}$ | Traffic counts for right-turn movement of approach $a$ of intersection $i$ at the time interval $t$ |

## 2.2. Data Encoding

To make road type, left-turn type, and datetime accessible by the model, data are encoded as follows:

1) **Road type**: The major road is encoded as 1 and minor road is encoded as 2. Road type information for approach $a$ of intersection $i$ is denoted as $r^{a,i}$.
2) **Left-turn type**: Considering there are different types of left-turn phases, permissive-only left-turn is encoded as 1, protected-permissive left-turn is encoded as 2, and protected-only left-turn is encoded as 3. Left-turn type information for approach $a$ of intersection $i$ is denoted as $l^{a,i}$.
3) **Minute-of-hour and hour-of-day**: Each hour has four 15-minute intervals, therefore, the minute-of-hour (MOH) takes values from 1 to 4 to represent each 15-minute interval. Hour-of-day (HOD) is represented by subsequent numbers, starting from 0 to 23 (0 representing midnight, and 23 representing 11 pm). MOH information and HOD information for the time interval $t$ are denoted as $h_{MOH}^{t}$ and $h_{HOD}^{t}$.

## 2.3. Problem Formulation

The purpose of this study is to build a framework to estimate TMCs for intersections. This goal can be achieved through training models on a known intersection (henceforth referred to as "source domain") and transferring the well-trained models to estimate TMCs for a new intersection (henceforth referred to as "target domain"). As shown in Eq. (1), Eq. (2), and Eq. (3), by associating traffic variables for a known intersection, the learned function $F_1(\cdot)$, $F_2(\cdot)$, and $F_3(\cdot)$, can be used for left-turn, through, and right-turn movement counts estimation on a new intersection.

$$F_1\left(\begin{bmatrix} o_{TM}^{a,i,t}, d_{TM}^{a,i,t}, g_{TM}^{a,i,t}, c_{TM}^{a,i,t}, m_{TM}^{a,i,t}, s_{TM}^{a,i,t}, o_{LM}^{a,i,t}, d_{LM}^{a,i,t}, g_{LM}^{a,i,t}, c_{LM}^{a,i,t}, m_{LM}^{a,i,t}, \\ s_{LM}^{a,i,t}, p_{LM}^{a,i,t}, l_{SL}^{a,i}, l_{EL}^{a,i}, l_{TL}^{a,i}, l_{ER}^{a,i}, l_{SR}^{a,i}, e_{POIE}^{i}, e_{POIC}^{i}, r^{a,i}, l^{a,i}, h_{MOH}^{t}, h_{HOD}^{t} \end{bmatrix}\right) = [v_{LM}^{a,i,t}] \quad (1)$$

$$F_2\left(\begin{bmatrix} o_{TM}^{a,i,t}, d_{TM}^{a,i,t}, g_{TM}^{a,i,t}, c_{TM}^{a,i,t}, m_{TM}^{a,i,t}, s_{TM}^{a,i,t}, o_{LM}^{a,i,t}, d_{LM}^{a,i,t}, g_{LM}^{a,i,t}, c_{LM}^{a,i,t}, m_{LM}^{a,i,t}, \\ s_{LM}^{a,i,t}, p_{LM}^{a,i,t}, l_{SL}^{a,i}, l_{EL}^{a,i}, l_{TL}^{a,i}, l_{ER}^{a,i}, l_{SR}^{a,i}, e_{POIE}^{i}, e_{POIC}^{i}, r^{a,i}, l^{a,i}, h_{MOH}^{t}, h_{HOD}^{t} \end{bmatrix}\right) = [v_{TM}^{a,i,t}] \quad (2)$$

$$F_3\left(\begin{bmatrix} o_{TM}^{a,i,t}, d_{TM}^{a,i,t}, g_{TM}^{a,i,t}, c_{TM}^{a,i,t}, m_{TM}^{a,i,t}, s_{TM}^{a,i,t}, o_{LM}^{a,i,t}, d_{LM}^{a,i,t}, g_{LM}^{a,i,t}, c_{LM}^{a,i,t}, m_{LM}^{a,i,t}, \\ s_{LM}^{a,i,t}, p_{LM}^{a,i,t}, l_{SL}^{a,i}, l_{EL}^{a,i}, l_{TL}^{a,i}, l_{ER}^{a,i}, l_{SR}^{a,i}, e_{POIE}^{i}, e_{POIC}^{i}, r^{a,i}, l^{a,i}, h_{MOH}^{t}, h_{HOD}^{t} \end{bmatrix}\right) = [v_{RM}^{a,i,t}] \quad (3)$$

Given source domain data $D_S = \{(x_{S_1}, y_{S_1}), \ldots, (x_{S_n}, y_{S_n})\}$, where $x_{S_i} \in \mathcal{X}_S$ is the data instance and $y_{S_i} \in \mathcal{Y}_S$ is the corresponding label. $F_1(\cdot)$, $F_2(\cdot)$, and $F_3(\cdot)$ can be generalized as $F_4(\cdot)$.

$$F_4([\mathcal{X}_S]) = [\mathcal{Y}_S] \quad (4)$$

Let target domain data $D_T = \{(x_{T_1}, y_{T_1}), \ldots, (x_{T_k}, y_{T_k})\}$, where $x_{T_i} \in \mathcal{X}_T$ is the data instance and $y_{T_i} \in \mathcal{Y}_T$ is the corresponding label. Assume $\mathcal{X}_T'$ and $\mathcal{Y}_T'$ are small portions of data from the target domain, $\mathcal{X}_S'$ and $\mathcal{Y}_S'$ are small portions of data from the source domain. The proposed framework employed in this study needs $\mathcal{X}_T'$ and $\mathcal{Y}_T'$ to be added to the model training process to fine-tune $F_4(\cdot)$ to acquire $F_5(\cdot)$. The proposed framework is able to reduce the weights of source instances that are dissimilar to the target instances and increase the weights of source instances that are similar to the target instances (Dai et al., 2007; Ma, Karimpour, et al., 2024). In this case, $F_5(\cdot)$ could have satisfactory performance for the target domain.

$$F_5([\mathcal{X}_S, \mathcal{X}_T']) = [\mathcal{Y}_S, \mathcal{Y}_T'] \quad (5)$$

Since $\mathcal{Y}_T'$ is not available in this study, $\mathcal{X}_S'$ and $\mathcal{Y}_S'$ are extracted from the source domain to substitute for $\mathcal{X}_T'$ and $\mathcal{Y}_T'$. While in practice, data scarcity is a common problem in TMC estimation. Generating synthetic data using GMM can be an effective approach to bridge the gap between domains and improve the performance of machine learning models in low-resource settings. Therefore, $\mathcal{X}_S'$ and $\mathcal{Y}_S'$ are identified using ITML and augmented by GMM, they become $\mathcal{X}_S''$ and $\mathcal{Y}_S''$, then they are substituted for $\mathcal{X}_T'$ and $\mathcal{Y}_T'$. In the end, $F_6(\cdot)$ is built for TMC estimation in this study.

$$F_6([\mathcal{X}_S, \mathcal{X}_S'']) = [\mathcal{Y}_S, \mathcal{Y}_S''] \quad (6)$$

## 3. METHODOLOGY

### 3.1. Research Framework

The main idea behind the proposed framework is to estimate TMCs for intersections. The proposed framework for estimating TMCs using DA consists of several key components, as depicted in the Figure 1. These components are designed to systematically process and utilize data from both source and target domains to estimate TMCs accurately.

In the first step, both source and target domain data undergo feature extraction processes to identify relevant traffic parameters. Feature extraction is a critical step as it transforms raw traffic data into structured information that can be utilized in subsequent steps. Let's assume the entire data set of the source and target domain intersections are denoted as $\mathbb{D}_S$ and $\mathbb{D}_T$, respectively; in this case, $\mathbb{D}_S = (\mathbb{X}_S, \mathbb{Y}_S)$ and $\mathbb{D}_T = (\mathbb{X}_T, \mathbb{Y}_T)$. Both $\mathbb{D}_S$ and $\mathbb{D}_T$ have two parts: data instances $\mathbb{X}_S$ and $\mathbb{X}_T$ as well as labels $\mathbb{Y}_S$ and $\mathbb{Y}_T$. Assume $\mathbb{X}_S$ and $\mathbb{X}_T$ can be represented as $n$ by $p$ matrices.

In the second step, feature selection techniques, such as Lasso Regression, is employed to identify the most significant features contributing to TMC patterns. This step helps in reducing the dimensionality of the data, enhancing model performance, and minimizing computational complexity. Effective feature selection ensures that the DA models focus on the most informative attributes. Recall that $D_S$ represents data selected from the source domain intersections and $D_T$ represents data selected from the target domain intersection. Similarly, $D_S = (\mathcal{X}_S, \mathcal{Y}_S)$ and $D_T = (\mathcal{X}_T, \mathcal{Y}_T)$. $\mathcal{X}_S$ and $\mathcal{X}_T$ contain all the variables extracted from the second step, assume the number of variables selected is $q$, then both $\mathcal{X}_S$ and $\mathcal{X}_T$ become $n$ by $q$ matrices.

With the important variables selected, an ITML algorithm is proposed to match each data instance in the target domain to a certain data instance of the source domain. The core of the ITML algorithm involves matching data instances in the source domain with those in the target domain based on their similarity. ITML leverages the selected features to find data instances with analogous traffic characteristics. This matching is crucial for ensuring that the knowledge transferred from the source domain is relevant and applicable to the target domain. Recall $\mathcal{X}_T'$ and $\mathcal{Y}_T'$ are small portions of data from the target domain, $\mathcal{X}_S'$ and $\mathcal{Y}_S'$ are small portions of data from the source domain. The proposed framework employed in this study needs $\mathcal{X}_T'$ and $\mathcal{Y}_T'$ to be added to the model training process to fine-tune the model. Since $\mathcal{Y}_T'$ is not available in this study, $\mathcal{X}_S'$ and $\mathcal{Y}_S'$ are extracted from the source domain to substitute for $\mathcal{X}_T'$ and $\mathcal{Y}_T'$. The labeled source data used for substitution needs to have the highest similarity to the data instances in the target domain. ITML algorithm is used to achieve this goal.

After $\mathcal{X}_S'$ and $\mathcal{Y}_S'$ are extracted from the source domain, data scarcity may arise as a significant problem when proceeding with the following steps. Generating synthetic data using GMM can be an effective approach for domain adaptation. GMM is a probabilistic model that represents the underlying distribution of a dataset as a mixture of multiple Gaussian distributions, each defined by its mean and covariance. By fitting a GMM to the available source domain data, it is possible to capture the complex patterns and variability within the dataset. Once the model is trained, synthetic samples can be generated by sampling from the learned mixture components, ensuring that the synthetic data closely resembles the statistical properties of the original dataset. Assuming the augmented data are $\mathcal{X}_S''$ and $\mathcal{Y}_S''$. This synthetic data can then be used to augment the target domain, helping to bridge the gap between domains and improve the performance of machine learning models in low-resource settings. GMM-based data generation is particularly useful when the source and target domains share similar feature spaces but differ in data distributions, as it allows for controlled and realistic data augmentation.

Once $\mathcal{X}_S'$ and $\mathcal{Y}_S'$ are identified using ITML and augmented by GMM, they become $\mathcal{X}_S''$ and $\mathcal{Y}_S''$, then they are substituted for $\mathcal{X}_T'$ and $\mathcal{Y}_T'$. This substitution process effectively creates a synthetic dataset for the target domain, enabling the use of existing data to infer traffic patterns where physical sensors are absent. In summary, target domain data substitution is conducted to extract $\mathcal{X}_S'$ and $\mathcal{Y}_S'$ from the source domain intersection to substitute for $\mathcal{X}_T'$ and $\mathcal{Y}_T'$.

Next, a DA model named GBBW is deployed to estimate TMCs for all the intersections in the target domain. In the model training process, labeled data instances that are similar to the $\mathcal{X}_S''$ and $\mathcal{Y}_S''$ are assigned higher weights, whereas those that are less similar to $\mathcal{X}_S''$ and $\mathcal{Y}_S''$ are given lower weights. By emphasizing labeled data instances with higher weights, the model prioritizes more relevant information, which in turn enhances the learning algorithm's capability to train a more accurate and effective regressor. The final output of the framework is the estimated TMCs. These predictions provide valuable insights for traffic management authorities, aiding in traffic signal optimization, congestion mitigation, and strategic planning.

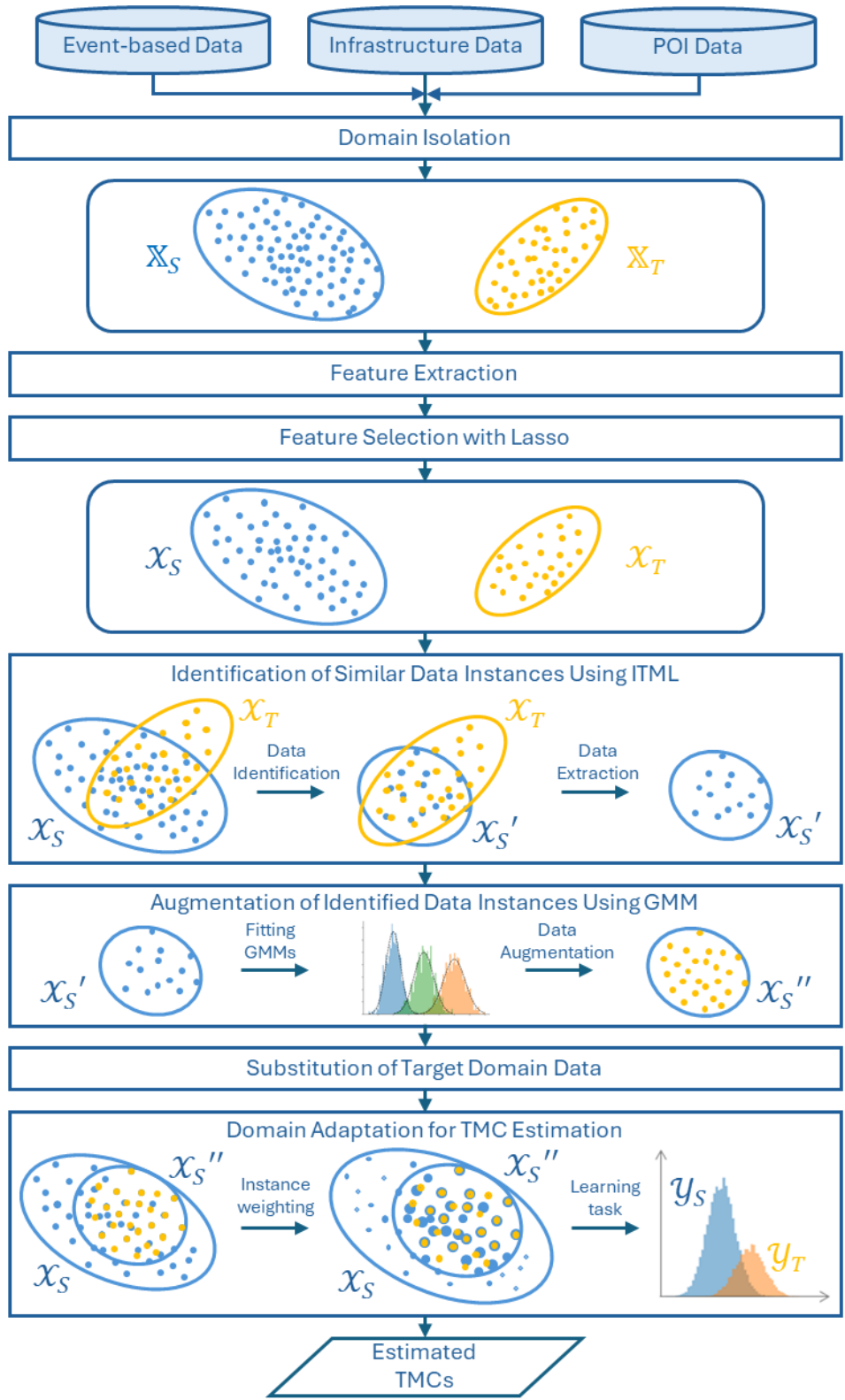

**Figure 1 Research framework**

## 3.2. Lasso Regression

In this study, Lasso regression is used for variable importance selection and spatial-temporal association. Lasso regression is a popular technique in the field of statistics and machine learning for both regularization and feature selection in linear regression models. It addresses some of the limitations of traditional linear regression by imposing a penalty on the absolute size of the regression coefficients. This penalty can lead to sparse solutions, where some coefficients are exactly zero, effectively performing variable selection (Tibshirani, 1996). When running Lasso regression, the independent variable is $\mathbb{X}_S$ and the dependent variable is $\mathbb{Y}_S$. Assume $\mathbb{X}_S$ can be represented as $n$ by $p$ matrice. $\mathcal{X}_S$ contains all the variables extracted based on Lasso regression, assuming the number of variables selected is $q$, then $\mathcal{X}_S$ becomes $n$ by $q$ matrix.

In a standard linear regression model, we aim to fit a linear relationship between the dependent variable $y$ and a set of independent variables $X = \{x_1, x_2, \ldots, x_p\}$. The model can be represented as:

$$y = \beta_0 + \beta_1 x_1 + \beta_2 x_2 + \cdots + \beta_p x_p + \epsilon \tag{7}$$

Where $y$ is the dependent variable, $\beta_0$ is the intercept, $\beta_i$ (for $i = 1, 2, \ldots, p$) are the regression coefficients, $x_i$ (for $i = 1, 2, \ldots, p$) are the independent variables, $\epsilon$ is the error term.

The goal of linear regression is to find the values of $\beta_0, \beta_1, \ldots, \beta_p$ that minimize the residual sum of squares (RSS):

$$\text{RSS} = \sum_{i=1}^{n}(y_i - \beta_0 - \sum_{j=1}^{p} x_{ij}\beta_j)^2 \tag{8}$$

Lasso regression modifies the objective function of linear regression by adding a regularization term that penalizes the absolute values of the regression coefficients. The objective function for Lasso regression is minimize:

$$\sum_{i=1}^{n}(y_i - \beta_0 - \sum_{j=1}^{p} x_{ij}\beta_j)^2 + \lambda \sum_{j=1}^{p}|\beta_j| \tag{9}$$

where $\lambda$ is a tuning parameter that controls the strength of the penalty. The term $\lambda \sum_{j=1}^{p}|\beta_j|$ is the L1 regularization term, which encourages sparsity in the coefficients $\beta_j$.

Lasso regression is a powerful tool for regularization and feature selection in linear models. By introducing an L1 penalty, it shrinks the regression coefficients and promotes sparsity, making it particularly useful in high-dimensional datasets. Its ability to perform variable selection and reduce overfitting makes it a valuable technique in both statistical analysis and machine learning applications.

## 3.3. Information-Theoretic Metric Learning

After choosing the important traffic variables using lasso regression, the next step is to find a matching function to map each data instance of the target domain to a corresponding data instance of the source domain. The objective is to find a labeled source data instance that has the highest similarity to each data instance in the target domain. Since $\mathcal{Y}_T'$ is not available in this study, $\mathcal{X}_S'$ and $\mathcal{Y}_S'$ are extracted from the source domain to substitute for $\mathcal{X}_T{}'$ and $\mathcal{Y}_T{}'$. ITML is employed to identify $\mathcal{X}_S{}'$ and $\mathcal{Y}_S{}'$. ITML customizes distance metrics to specific tasks by leveraging similarity constraints and prior knowledge, improving performance in tasks like classification, clustering, and retrieval. It ensures robustness through convex optimization, balances generalization with regularization, and has strong theoretical foundations rooted in information theory. ITML is scalable to high-dimensional data and adapts well to real-world applications, outperforming standard metrics like Euclidean distance (Davis et al., 2007). Combining $\mathcal{X}_S{}'$ and $\mathcal{X}_T{}'$ gives a set of $n$ points $\{x_1, x_2, \ldots, x_n\}$ in $\mathbb{R}^d$, we aim to find a positive definite matrix $A$ that parameterizes the Mahalanobis distance.

$$d_A(x_i, x_j) = (x_i - x_j)^T A(x_i - x_j) \tag{10}$$

Two points are considered similar if the Mahalanobis distance between them is less than or equal to a specified upper bound, i.e., $d_A(x_i, x_j) \leq u$, where $u$ is relatively small. Conversely, two points are dissimilar if $d_A(x_i, x_j) \geq l$, where $l$ is sufficiently large.

The Mahalanobis matrix $A$ is regularized to remain as close as possible to a predefined Mahalanobis distance function, parameterized by $A_0$. The measure of "closeness" between $A$ and $A_0$ is quantified using a natural information-theoretic approach.

Given a Mahalanobis distance parameterized by $A$, its corresponding multivariate Gaussian is expressed as $p(x; A) = \frac{1}{Z} \exp\left(-\frac{1}{2} d_A(x, \mu)\right)$, where $Z$ is a normalizing constant and the Gaussian has mean $\mu$. Through this bijection, the distance between two Mahalanobis distance functions, parameterized by $A_0$ and $A$, is measured using the (differential) relative entropy between their corresponding multivariate Gaussians:

$$\mathrm{KL}(p(x; A_0) \| p(x; A)) = \int p(x; A_0) \log \frac{p(x; A_0)}{p(x; A)} dx \tag{11}$$

Given pairs of similar points $S$ and pairs of dissimilar points $D$, the problem of learning the distance metric is defined as follows

$$\min_A \mathrm{KL}(p(x; A_0) \| p(x; A)) \tag{12}$$

subject to

$$d_A(x_i, x_j) \leq u \quad (i, j) \in S,$$

$$d_A(x_i, x_j) \geq l \quad (i, j) \in D.$$

The LogDet divergence is a Bregman matrix divergence derived from the convex function $\phi(X) = -\log \det X$ defined over the cone of positive-definite matrices. For $n \times n$ matrices $A, A_0$, it is given by:

$$D_{\ell d}(A, A_0) = \mathrm{tr}(AA_0^{-1}) - \log \det (AA_0^{-1}) - n \tag{13}$$

The differential relative entropy between two multivariate Gaussians has been shown to be expressible as a convex combination of the Mahalanobis distance between their mean vectors and the LogDet divergence between their covariance matrices (Davis & Dhillon, 2006) Assuming the means of the Gaussians are identical, this simplifies to:

$$\mathrm{KL}(p(x; A_0) || p(x; A)) = \tfrac{1}{2} D_{\ell\mathrm{d}}(A_0^{-1}, A^{-1}) = \tfrac{1}{2} D_{\ell\mathrm{d}}(A, A_0) \tag{14}$$

The result can be further extended to exhibit invariance under any invertible linear transformation $S$, since

$$D_{\ell\mathrm{d}}(S^T A S, S^T B S) = D_{\ell\mathrm{d}}(A, B) \tag{15}$$

The equivalence in Eq. (14) can be utilized to reformulate the distance metric learning problem Eq. (12) as the following LogDet optimization problem:

$$\min_{A \geq 0} D_{\ell\mathrm{d}}(A, A_0) \tag{16}$$

subject to

$$\mathrm{tr}(A(x_i - x_j)(x_i - x_j)^T) \leq u \quad (i,j) \in S,$$

$$\mathrm{tr}(A(x_i - x_j)(x_i - x_j)^T) \geq l \quad (i,j) \in D.$$

In certain situations, a feasible solution to Eq. (16) may not exist. To avoid such scenarios, slack variables are introduced into the formulation to ensure the existence of a feasible matrix $A$. Let $c(i,j)$ represent the index of the $(i,j)$-th constraint, and let $\xi$ be a vector of slack variables initialized to $\xi_0$, where its components are set to $u$ for similarity constraints and $l$ for dissimilarity constraints. With this, the optimization problem can be reformulated as follows:

$$\min_{A \geq 0, \xi} D_{\ell\mathrm{d}}(A, A_0) + \gamma \cdot D_{\ell\mathrm{d}}(diag(\xi), diag(\xi_0)) \tag{17}$$

subject to

$$\mathrm{tr}(A(x_i - x_j)(x_i - x_j)^T) \leq \xi_{c(i,j)} \quad (i,j) \in S,$$

$$\mathrm{tr}(A(x_i - x_j)(x_i - x_j)^T) \geq \xi_{c(i,j)} \quad (i,j) \in D.$$

The parameter $\gamma$ regulates the tradeoff between fulfilling the constraints and minimizing $D_{\ell\mathrm{d}}(A, A_0)$.

The optimization problem of Eq. (17) is solved by extending the methods from (Kulis et al., 2006). The optimization approach underlying the algorithm involves repeatedly computing Bregman projections, which are projections of the current solution onto a single constraint. This projection is carried out using the update:

$$A_{t+1} = A_t + \beta A_t (x_i - x_j)(x_i - x_j)^T A_t \tag{18}$$

Where $x_i$ and $x_j$ represent the constrained data points, and $\beta$ is the projection parameter (Lagrange multiplier corresponding to the constraint) computed by the algorithm. The resulting algorithm is outlined below.

**Input** $X$: input $d \times n$ matrix, $S$: set of similar pairs, $D$: set of dissimilar pairs, $u, l$: distance thresholds, $A_0$: input Mahalanobis matrix, $\gamma$: slack parameter, $c$: constraint index function

$A \leftarrow A_0, \lambda_{ij} \leftarrow 0 \; \forall \; i,j$
$\xi_{c(i,j)} \leftarrow u$ for $(i,j) \in S$; otherwise $\xi_{c(i,j)} \leftarrow l$
Repeat until convergence:
    1. Pick a constraint $(i,j) \in S$ or $(i,j) \in D$
    2. $p \leftarrow (x_i - x_j)^T A (x_i - x_j)$
    3. $\delta \leftarrow 1$ if $(i,j) \in S$, $-1$ otherwise
    4. $\alpha \leftarrow \min(\lambda_{ij}, \frac{\delta}{2}(\frac{1}{p} - \frac{\gamma}{\xi_{c(i,j)}}))$
    5. $\beta \leftarrow \frac{\delta \alpha}{1 - \delta \alpha p}$
    6. $\xi_{c(i,j)} \leftarrow \frac{\gamma \xi_{c(i,j)}}{\gamma + \delta \alpha \xi_{c(i,j)}}$
    7. $\lambda_{ij} \leftarrow \lambda_{ij} - \alpha$
    8. $A = A + \beta A (x_i - x_j)(x_i - x_j)^T A$
End Repeat
**Output** Mahalanobis matrix $A$

### 3.4. Gaussian Mixture Model

GMM is a probabilistic model that represents a dataset as a mixture of multiple Gaussian distributions. It is commonly used for clustering, density estimation, and modeling data with subpopulations. A GMM models the probability distribution of a dataset $X = \{x_1, x_2, \ldots, x_N\}$, where $x_i \in \mathbb{R}^d$, as a weighted sum of $K$ Gaussian components. The probability density function (PDF) of the GMM is:

$$p(x) = \sum_{k=1}^{K} \pi_k \mathcal{N}(x \mid \mu_k, \Sigma_k) \tag{19}$$

Where:

$K$: Number of Gaussian components.

$\pi_k$: Mixing coefficient for the $k$-th Gaussian component, such that $\sum_{k=1}^{K} \pi_k = 1$ and $\pi_k \geq 0$.

$\mathcal{N}(x \mid \mu_k, \Sigma_k)$: Multivariate Gaussian distribution for the $k$-th component with mean $\mu_k \in \mathbb{R}^d$ and covariance matrix $\Sigma_k \in \mathbb{R}^{d \times d}$.

The Gaussian PDF is defined as:

$$\mathcal{N}(x \mid \mu_k, \Sigma_k) = \frac{1}{(2\pi)^{\frac{d}{2}} |\Sigma_k|^{\frac{1}{2}}} \exp(-\frac{1}{2}(x - \mu_k)^T \Sigma_k^{-1} (x - \mu_k)) \tag{20}$$

### 3.4.1. Gaussian Mixture Model for Data Augmentation

In practice, data scarcity is a common problem in TMC estimation. Addressing this issue becomes a critical challenge in the subsequent steps. One effective solution is synthetic data generation using GMM. GMM is a probabilistic model that represents a dataset's distribution as a mixture of Gaussian components, each defined by its mean and covariance (X. Chen et al., 2025). By fitting a GMM to the source domain data, the model captures the dataset's patterns and variability. Once trained, synthetic samples can be generated by sampling from the mixture components, ensuring that the synthetic data reflects the statistical properties of the original dataset. This augmented data can then be used to support the target domain, reducing the domain gap and improving the performance of machine learning models in low-resource scenarios. GMM-based data augmentation is particularly effective when the source and target domains share similar feature spaces but differ in data distributions. Assuming $\mathcal{X}_S'$ and $\mathcal{Y}_S'$ are extracted from the source domain with $N$ samples ( $d$-dimensional, $d = q + 1$ ), below is the pseudocode for data augmentation using GMM.

---

**Input** Original dataset $X = \{x_1, x_2, \ldots, x_N\}$ with $N$ samples ($d$-dimensional), number of Gaussian components $K$, number of synthetic samples to generate $M$

**Step 1**: Fit GMM to the original dataset
Initialize GMM with $K$ components
Fit GMM parameters ($\pi_k, \mu_k, \Sigma_k$) using Expectation-Maximization (EM) algorithm:
    Repeat until convergence:
        // E-step: Calculate responsibilities
        For $i = 1$ to $N$:
            For $k = 1$ to $K$:
$$\gamma_{ik} = \frac{\pi_k \mathcal{N}(x_i \mid \mu_k, \Sigma_k)}{\sum_{j=1}^{K} \pi_j \mathcal{N}(x_i \mid \mu_j, \Sigma_j)}$$
            End for
        End for
        // M-step: Update GMM parameters
        For $k = 1$ to $K$:
$$\pi_k = \frac{1}{N} \sum_{i=1}^{N} \gamma_{ik}$$
$$\mu_k = \frac{\sum_{i=1}^{N} \gamma_{ik} x_i}{\sum_{i=1}^{N} \gamma_{ik}}$$
$$\Sigma_k = \frac{\sum_{i=1}^{N} \gamma_{ik}(x_i - \mu_k)(x_i - \mu_k)^T}{\sum_{i=1}^{N} \gamma_{ik}}$$
        End for
    End Repeat

**Step 2**: Generate synthetic data
Initialize $X_{\text{synthetic}} = \{\}$
For $j = 1$ to $M$:
    1. Randomly select a Gaussian component $k$ from $\{1, 2, \ldots, k\}$ based on $\pi_k$
    2. Sample synthetic point $x_{new}$ from the selected Gaussian: $x_{new} \sim \mathcal{N}(\mu_k, \Sigma_k)$
    3. Add new synthetic point to the set $X_{\text{synthetic}} = X_{\text{synthetic}} \cup \{x_{new}\}$
End for

**Step 3**: Combine original and synthetic data
$$X_{\text{augmented}} = X \cup X_{\text{synthetic}}$$

**Output** $X_{\text{augmented}}$

---

### 3.5. Target Domain Data Substitution

Recall source domain data $D_S = \{(x_{S_1}, y_{S_1}), \ldots, (x_{S_n}, y_{S_n})\}$, where $x_{S_i} \in \mathcal{X}_S$ is the data instance and $y_{S_i} \in \mathcal{Y}_S$ is the corresponding label. Target domain data is denoted as $D_T = \{(x_{T_1}, y_{T_1}), \ldots, (x_{T_k}, y_{T_k})\}$, where $x_{T_i} \in \mathcal{X}_T$ is the data instance and $y_{T_i} \in \mathcal{Y}_T$ is the corresponding label. When applying DA to induce a predictive model, some labeled data in the target domain are required. However, in practice, labeled data in the target domain are not available. This study chooses an alternative approach that uses labeled source domain data as a substitute for labeled target domain data. The labeled source data used for substitution needs to have the highest similarity to the data instances in the target domain.

Specifically, once $\mathcal{X}_S'$ and $\mathcal{Y}_S'$ are identified using ITML and augmented by GMM, they become $\mathcal{X}_S''$ and $\mathcal{Y}_S''$, then they are substituted for $\mathcal{X}_T'$ and $\mathcal{Y}_T'$. This substitution process effectively creates a synthetic dataset for the target domain, enabling the use of existing data to infer traffic patterns where physical sensors are absent.

### 3.6. Gradient Boosting

Gradient boosting is used in supervised learning to find a function $\hat{F}(x)$ that best predicts an output variable $y$ from input variables $x$. This is achieved by introducing a loss function $L(y, F(x))$, and minimizing its expected value:

$$\hat{F} = \arg\min_F \mathbb{E}_{x,y}[L(y, F(x))] \tag{21}$$

The method approximates $\hat{F}(x)$ as a sum of $M$ simpler functions $h_m(x)$ from some class $\mathcal{H}$, called base or weak learners:

$$\hat{F}(x) = \sum_{m=1}^{M} \gamma_m h_m(x) + \text{const.} \tag{22}$$

Given a training set of $\{(x_1, y_1), \ldots, (x_n, y_n)\}$, the algorithm aims to minimize the empirical risk (average loss on the training set). It starts with a constant function $F_0(x)$ and builds the model incrementally:

$$F_0(x) = \arg\min_\gamma \sum_{i=1}^{n} L(y_i, \gamma) \tag{23}$$

$$F_m(x) = F_{m-1}(x) + (\arg\min_{h_m \in \mathcal{H}} [\sum_{i=1}^{n} L(y_i, F_{m-1}(x_i) + h_m(x_i))])(x) \tag{24}$$

Since finding the optimal $h_m$ at each step is computationally infeasible, a simplified approach is used. The algorithm applies steepest descent, moving a small amount $\gamma$ in the negative gradient direction of the loss function:

$$F_m(x) = F_{m-1}(x) - \gamma \sum_{i=1}^{n} \nabla_{F_{m-1}} L(y_i, F_{m-1}(x_i)) \tag{25}$$

This process continues for $M$ iterations, gradually improving the model's predictive power by combining multiple weak learners into a strong predictor.

### 3.6.1. Gradient Boosting with Balanced Weighting

Balanced Weighting is an instance-based domain adaptation technique that assigns weights to samples to account for data distribution differences between source and target domains. It balances the source distribution by maximizing the weights of samples similar to the target distribution and minimizing irrelevant ones. This improves model performance on target tasks, ensures better generalization, and effectively handles domain shifts (De Mathelin et al., 2022b). This study proposes GBBW as an extension of Gradient Boosting by integrating the Balanced Weighting technique for domain adaptation. This approach builds on the strengths of Gradient Boosting, a powerful ensemble learning method, and incorporates Balanced Weighting to address distribution differences between source and target domains. By emphasizing samples from the source domain that align closely with the target distribution, it effectively handles domain shifts while leveraging gradient boosting's strong predictive performance. This method enhances model generalization to target tasks and improves accuracy in scenarios with distribution differences between source and target domains. In the GBBW, the ratio parameter, $\alpha$, controls the balance between the influence of source and target data in the loss function. This approach enables the model to effectively leverage information from both domains, enhancing its generalization capability across diverse datasets. By dynamically adjusting the contributions of source and target data, the method facilitates domain adaptation by weighting the importance of each domain based on the value of $\alpha$.

During the initialization step, both source and target data are incorporated, with weights of $1 - \alpha$ and $\alpha$, respectively. Pseudo-residuals are calculated separately for the source and target data, and the base learner is trained on a combined set of pseudo-residuals from both domains, weighted by $1 - \alpha$ for source samples and $\alpha$ for target samples. The multiplier computation further accounts for losses from both source and target data, applying weights of $1 - \alpha$ and $\alpha$ accordingly. This approach allows the Gradient Boosting algorithm to learn from both source and target domains, effectively balancing their contributions through the $\alpha$ parameter. By assigning appropriate importance to each domain, it significantly aids in domain adaptation tasks during the learning process. To simplify the notation, the source domain data is denoted as $\{(x_i, y_i)\}_{i=1}^{n_1}$ (source), which is $D_S$ in this study. The target domain data is denoted as $\{(x_j, y_j)\}_{j=1}^{n_2}$ (target), which corresponds to $\mathcal{X}_S''$ and $\mathcal{Y}_S''$ that are identified using ITML and augmented by GMM.

---

**Input** Two training sets $\{(x_i, y_i)\}_{i=1}^{n_1}$ (source) and $\{(x_j, y_j)\}_{j=1}^{n_2}$ (target), a differentiable loss function $L(y, F(x))$, number of iterations $M$, and a weighting parameter $\alpha$.
**Initialize model with a constant value:**

$$F_0(x) = \arg\min_{\gamma} ((1 - \alpha) \sum_{i=1}^{n_1} L(y_i, \gamma) + \alpha \sum_{j=1}^{n_2} L(y_j, \gamma))$$

**For** $m = 1$ to $M$:
  1. Computer pseudo-residuals for both training sets:

$$r_{im} = -\left[\frac{\partial L(y_i, F(x_i))}{\partial F(x_i)}\right]_{F(x)=F_{m-1}(x)} \quad \text{for } i = 1, \dots, n_1$$

$$r_{jm} = -\left[\frac{\partial L(y_j, F(x_j))}{\partial F(x_j)}\right]_{F(x)=F_{m-1}(x)} \quad \text{for } j = 1, \dots, n_2$$

  2. Fit a base learner $h_m(x)$ to pseudo-residuals:

Train it using the combined training set $\{(x_i, r_{im})\}_{i=1}^{n_1} \cup \{(x_j, r_{jm})\}_{j=1}^{n_2}$

3. Compute multiplier $\gamma_m$ by solving the following optimization problem:

$$\gamma_m = \arg\min_\gamma ((1-\alpha)\sum_{i=1}^{n_1} L(y_i, F_{m-1}(x_i) + \gamma h_m(x_i)) + \alpha \sum_{j=1}^{n_2} L(y_j, F_{m-1}(x_j) + \gamma h_m(x_j)))$$

4. Update the model:

$$F_m(x) = F_{m-1}(x) + \gamma_m h_m(x)$$

**Output** $F_M(x)$

## 4. EXPERIMENTS

### 4.1. Data Description

The data utilized in this study was gathered from the eastern Pima County region and encompasses three distinct categories: event-based data, intersection infrastructure data, and POI data.

### 4.1.1. Event-based Data

The event-based data in this study, collected via the MaxView system—an Advanced Traffic Management System (ATMS)—comprises a series of real-time events. These include detector actuation events, signal change events, pedestrian-related events, and controller communication events. For the purposes of this research, three specific types of event datasets are utilized: detector actuation events, signal change events, and communication events. These datasets are instrumental in deriving event information pertinent to left-turn and through movements, as detailed in Table 1.

### 4.1.2. Infrastructure Data

In the proposed framework, intersection infrastructure data is manually gathered from Google Maps and encompasses various elements. This data includes the classification of roads (major or minor), the number of left-turn lanes, shared left-turn lanes, through lanes, right-turn lanes, and shared right-turn lanes.

### 4.1.3. POI Data

POI data serves as an indicator of urban land use context and economic activities that influence traffic attraction and generation. In this study, the POI data was sourced from the Pima Association of Governments (PAG) employment database. The authenticity of the business presence and the quality of data were validated using the Google Places API and sample reviews (Kramer et al., 2022; Noh et al., 2019). Analysis within this study includes extracting counts of various categories and the number of employees from raw POI data within a 400m radius of an intersection. These category counts provide insights into the diversity of POI and land use characteristics around intersections, while the scale of each POI category is indicated by the employee counts. Together, these metrics are crucial for understanding the traveler characteristics and traffic patterns at nearby intersections.

## 4.2. Experiment Design

### 4.2.1. Data Sample

The TMC data provided by PAG was an aggregate of 15-minute interval TMC data during peak hours (7:00-9:00 AM, 4:00-6:00 PM) from 2016 to 2020. This data was typically gathered by consulting firms employing a combination of manual data collection methods and individual sensors. On average, data from one to four days per intersection were utilized for model training and validation. Figure 2 presents the layout of the study intersections located in Tucson, Arizona. A total of 30 intersections were randomly selected for inclusion in this study.

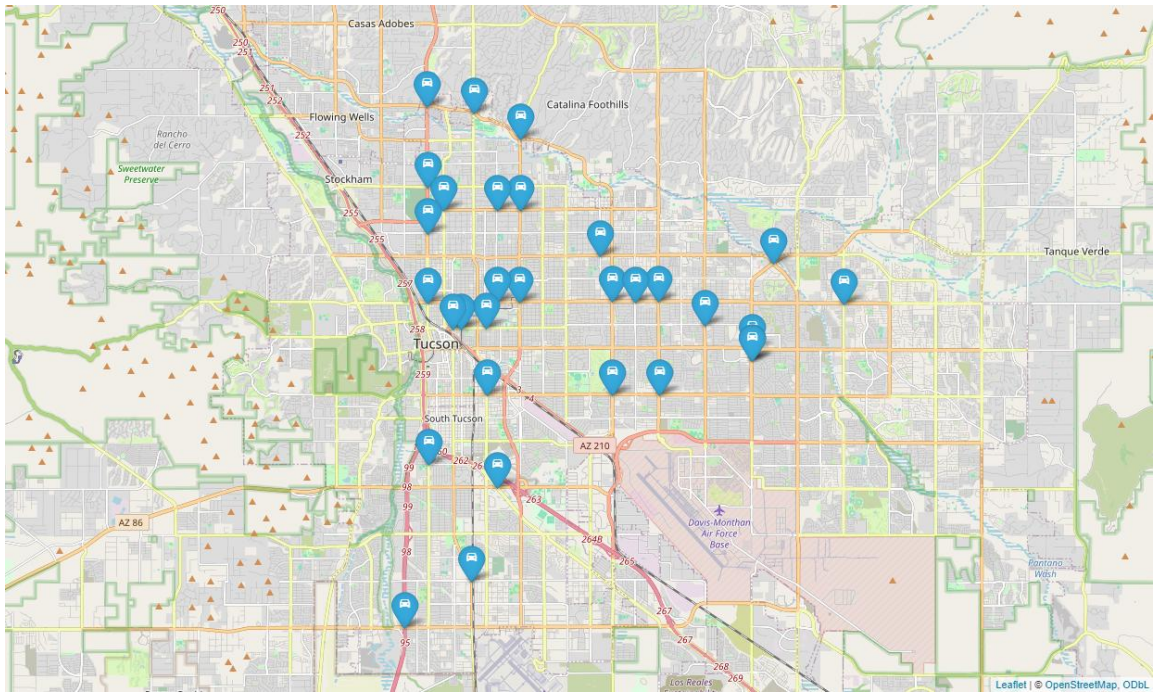

**Figure 2 The layout of the study intersections**

Figure 3 illustrates the distribution of data samples across various Intersection IDs. The x-axis represents the Intersection IDs, ranging from 6 to 728, while the y-axis shows the corresponding data sample counts. The data reveals significant variation in counts among the intersections, with some IDs, such as 551, having the highest count of approximately 120, while others, like 494 and 728, have minimal counts close to 5. This visualization highlights the uneven distribution of counts across the intersections. In total, 1,247 data samples were collected for these 30 intersections.

During the model training process, a dataset comprising 30 distinct intersections was utilized. In each iteration, one intersection was designated as the target domain, while the remaining 29 intersections acted as the source domain. The model was executed 30 times, with each iteration focusing on a different target domain, allowing for a comprehensive evaluation. Performance metrics were then calculated for each run to thoroughly assess the model's effectiveness and robustness across all possible scenarios, ensuring a reliable and accurate evaluation of its generalization capabilities.

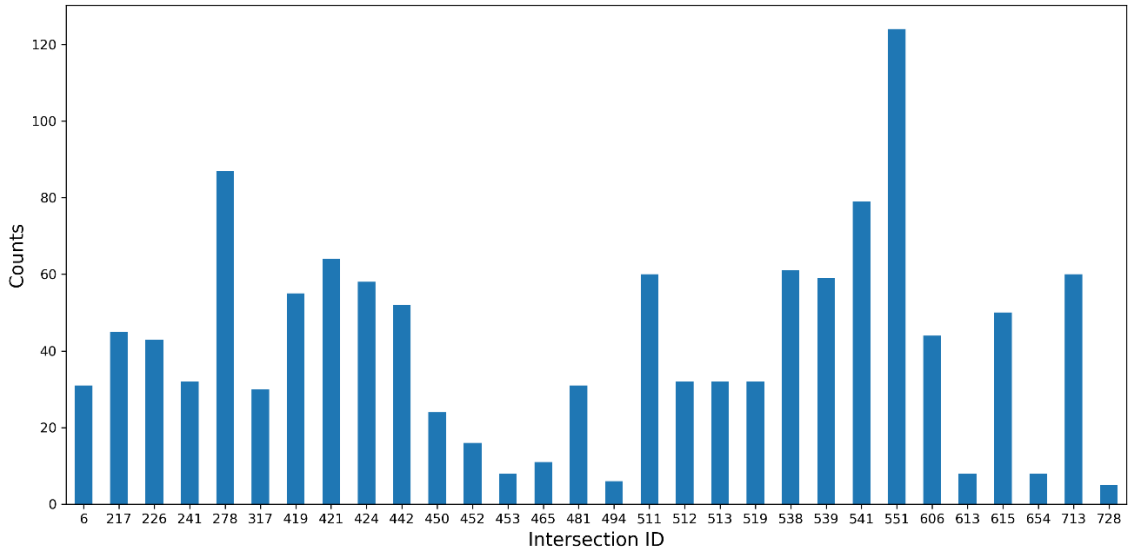

**Figure 3 Data sample for each Intersection ID**

**4.2.2. Baseline Models**

Seven state-of-the-art machine learning regression models were selected as baseline models to evaluate the feasibility of the proposed framework for estimating TMCs. These models include K-Nearest Neighbors (KNN), Support Vector Regression (SVR), Multi-Layer Perceptron (MLP), Adaptive Boosting (AdaBoost), eXtreme Gradient Boosting (XGBoost), Categorical Boosting (CatBoost), and Light Gradient Boosting Machine (LightGBM). Additionally, four state-of-the-art instance-based domain adaptation methods were included as baseline models for comparison: Importance Weighting Classifier (IWC), Importance Weighting Network (IWN), Relative Unconstrained Least-Squares Importance Fitting (RULSIF), and Weighting Adversarial Neural Network (WANN).

1) KNN(Cover & Hart, 1967): KNN is a simple machine learning algorithm that classifies data points based on the majority class or average of their k closest neighbors in the feature space. It is non-parametric, meaning it makes no assumptions about the underlying data distribution.
2) SVR(Drucker et al., 1996): SVR is a machine learning algorithm based on Support Vector Machines that is used for regression tasks. It aims to find a function that best fits the data while allowing a margin of tolerance (epsilon) around the predicted values to balance complexity and generalization.
3) MLP(Rumelhart et al., 1986): MLP is a type of artificial neural network composed of multiple layers of neurons, including an input layer, one or more hidden layers, and an output layer. It uses backpropagation and activation functions to learn complex patterns in data, making it effective for classification and regression tasks.

4) AdaBoost(Freund & Schapire, 1995): AdaBoost is an ensemble learning algorithm that combines multiple weak classifiers to create a strong classifier by iteratively adjusting the weights of incorrectly classified instances. It focuses more on hard-to-classify data points, improving model accuracy and robustness.
5) XGBoost(T. Chen & Guestrin, 2016): XGBoost is a highly efficient, scalable machine learning algorithm that uses gradient boosting techniques to build strong predictive models by combining multiple weak learners. It is known for its speed, performance, and ability to handle large datasets with high accuracy.
6) CatBoost(Prokhorenkova et al., 2018): CatBoost is a gradient boosting algorithm designed to handle categorical features efficiently without the need for extensive preprocessing. It leverages ordered boosting and other advanced techniques to achieve high performance, particularly in tasks involving heterogeneous data.
7) LightGBM(Ke et al., 2017): LightGBM is a fast, distributed gradient boosting framework that uses a histogram-based method to speed up training while maintaining high accuracy. It is optimized for large datasets and offers efficient handling of categorical features.
8) IWC(Bickel et al., 2007): Importance weighting is determined using the output of a domain classifier that distinguishes between source and target data.
9) IWN(De Mathelin et al., 2022a): The objective of IWN is to adjust the weights of source instances to minimize the Maximum Mean Discrepancy, ensuring closer alignment between the reweighted source distribution and the target distribution.
10) RULSIF(Yamada et al., 2011): The algorithm aims to correct discrepancies between the input distributions of the source and target domains. This is achieved by determining an optimal reweighting of source instances that minimizes the relative Pearson divergence between the source and target distributions.
11) WANN(De Mathelin et al., 2021): The goal of WANN is to determine a reweighting of source instances that compensates for distributional shifts between the source and target domains. This is accomplished by minimizing the Y-discrepancy distance between their distributions.

To build effective estimators for the target domain by leveraging information from the source domain, we integrated a metric learning approach with each of these domain adaptation methods, transforming them into supervised domain adaptation methods. The four resulting supervised domain adaptation methods are ITML-IWC, ITML-IWN, ITML-RULSIF, and ITML-WANN.

Considering the substantial influence of hyperparameter settings on the accuracy of each model's predictions, a grid search method was employed to identify the optimal hyperparameter values for both the baseline models and the proposed approach. The experiments were conducted on a platform equipped with an NVIDIA GeForce RTX 3080 GPU, a 12th Gen Intel(R) Core(TM) i7-12700KF CPU, and Python version 3.10.9.

### 4.2.3. Measurements of Effectiveness

Mean Absolute Error (MAE) and Root Mean Square Error (RMSE) are two common criteria used to evaluate and compare prediction methods (Luo et al., 2022). MAE is used to measure the overall errors of the estimation results and RMSE is used to quantify the stability of the estimation results (W. Zhang et al., 2022). These two criteria are employed as performance metrics for comparison in this study and are defined below:

$$\text{MAE} = \frac{1}{N}\sum_{k=1}^{N}|y(k) - \hat{y}(k)| \tag{26}$$

$$\text{RMSE} = \sqrt{\frac{1}{N}\sum_{k=1}^{N}(y(k) - \hat{y}(k))^2} \tag{27}$$

where $y(k)$ is the actual value at time $k$ and $\hat{y}(k)$ is the corresponding predicted value. $N$ is the size of the testing data set (total number of time intervals).

### 4.3. Lasso Regression Results

Lasso regression ranks the variables that have the highest impact on TMCs. In addition, it provides an approach to better understand the interaction between TMCs, detector data, lane configuration, signal timing, road characteristics, and temporal factors. To better identify the variables, their names are provided.

**Table 2 Lasso Regression Results**

| Variables | Left turn | Through | Right turn |
|---|---|---|---|
| Through movement detector occupancy time | -0.16 | 55.53 | 7.09 |
| Through movement detector trigger counts | 0.26 | 13.92 | -2.95 |
| Through movement green time duration | 0 | 19.15 | 0.21 |
| Through movement cycle counts | 2.04 | -0.47 | -0.68 |
| Through movement average of time differences between each pair of consecutive detections | 1.19 | 5.49 | 0 |
| Through movement standard deviation of time differences between each pair of consecutive detections | 0 | 14.67 | -1.31 |
| Left-turn movement detector occupancy time | 14.51 | 4.04 | -0.54 |
| Left-turn movement detector trigger counts | 0.43 | 0.69 | 2.40 |
| Left-turn movement green time duration | 16.18 | -5.18 | 3.39 |
| Left-turn movement cycle counts | -9.94 | 13.13 | 0.52 |
| Left-turn movement average of time differences between each pair of consecutive detections | -0.20 | 1.59 | 1.94 |
| Left-turn movement standard deviation of time differences between each pair of consecutive detections | 0 | 0.05 | 1.52 |
| Left-turn movement permissive green time | 5.13 | 2.25 | -4.15 |
| Number of shared left turn lanes | 2.50 | -5.41 | 1.29 |
| Number of exclusive left turn lanes | 10.65 | 0 | 7.81 |
| Number of through lanes | 1.98 | 29.58 | -0.88 |
| Number of exclusive right turn lanes | 1.31 | 14.51 | 9.58 |
| Number of shared right turn lanes | 0 | 0 | 0 |
| Number of employees of all POI | 3.21 | 15.29 | 0.78 |
| POI categories count | -4.68 | -4.35 | 0 |
| Road type | 1.78 | -57.63 | -8.10 |
| Left-turn type | 0 | -1.16 | 0 |
| Direction | -0.97 | -6.07 | -0.63 |
| Minute-of-hour | 0.42 | 2.27 | 0.24 |
| Hour-of-day | -0.38 | 3.60 | 1.01 |

Based on Table 2, several key variables significantly influence traffic movement counts (TMCs) for left-turn, through, and right-turn movements. The analysis highlights the importance of detector data, lane configurations, signal timing, road characteristics, and temporal factors in estimating traffic counts.

For through movements, through movement detector occupancy time has the strongest positive impact (55.53), indicating that higher detector occupancy correlates with increased through movement counts. Similarly, the number of through lanes (29.58) has a substantial positive influence, suggesting that intersections with more through lanes experience higher through traffic volumes. Other factors, such as through movement green time duration (19.15) and through movement detector trigger counts (13.92), further emphasize the importance of signal timing and detector activity in estimating through movement traffic. On the other hand, road type (-57.63) has a strong negative impact, indicating that minor roads are less conducive to high through traffic volumes, likely due to design or regulatory constraints.

For left-turn movements, left-turn movement green time duration (16.18) and the number of exclusive left-turn lanes (10.65) are the most influential factors. Longer green times and dedicated left-turn lanes positively correlate with higher left-turn traffic. Additionally, left-turn movement detector occupancy time (14.51) plays a significant role, further highlighting the importance of detector data. However, left-turn movement cycle counts (-9.94) negatively impact left-turn counts, suggesting that higher cycle counts may reduce left-turn traffic, possibly due to increased delays or inefficiencies.

Right-turn movements are strongly influenced by the number of exclusive right-turn lanes (9.58), which positively impacts right-turn counts, indicating that dedicated infrastructure facilitates higher right-turn traffic. Through movement detector occupancy time (7.09) also positively correlates with right-turn counts, likely reflecting increased overall traffic at the intersection. Interestingly, left-turn movement green time duration (3.39) positively influences right-turn counts, possibly due to coordinated signal timings that benefit multiple movements. Similar to through movements, road type (-8.10) negatively affects right-turn counts, underscoring the role of road design and surrounding conditions.

Other influential factors include temporal variables such as minute-of-hour (0.42 for left-turn, 2.27 for through, and 0.24 for right-turn) and hour-of-day (3.60 for through), which have varying impacts across movements, reflecting the influence of daily traffic patterns. The number of employees of all POI (3.21 for left-turn, 15.29 for through, and 0.78 for right-turn) positively affects traffic counts, particularly for through movements, likely due to increased activity in areas with more employees. Conversely, POI categories count has a negative impact across all movements (-4.68 for left-turn, -4.35 for through, and 0 for right-turn), suggesting that a higher diversity of POIs may reduce traffic volumes, possibly due to dispersed traffic patterns.

In conclusion, the findings underscore the complexity of traffic flow at intersections and the necessity of considering multiple variables to accurately estimate traffic counts. Detector data, lane configurations, signal timing, road characteristics, and temporal factors all play critical roles in understanding and managing traffic at intersections. The analysis highlights the importance of tailored infrastructure and signal timing to optimize traffic flow for different movement types.

**4.4. TMCs Estimation Results**

The inputs for both the baseline models and the proposed framework were the variables selected by the lasso regression, and the outputs were TMCs with 15-minute intervals. In the model training process, each model was tested using Lasso regression-selected variables, with one intersection being tested at a time and the remaining 29 used for training. This process was repeated 30 times to ensure all intersections were tested. The models evaluated include KNN, SVR, MLP, AdaBoost, XGBoost, CatBoost, LightGBM, ITML-IWC, ITML-IWN, ITML-RULSIF, ITML-WANN, ITML-GBBW, and ITMLGMM-GBBW. Two measures of effectiveness, MAE and RMSE were used for performance evaluation.

Table 3 presents a comparison of estimation performance, measured in terms of Mean Absolute Error (MAE), between seven state-of-the-art regression models, five domain adaptation models, and the proposed framework (ITMLGMM-GBBW) for estimating TMCs across three movement types: left-turn, through, and right-turn. The values represent the average MAE across 30 intersections for each movement type.

The proposed model, ITMLGMM-GBBW, achieved the lowest MAE values among all models, with 12.29 for left-turn, 34.39 for through, and 16.17 for right-turn movements. These results demonstrate its superior performance in estimating TMCs at intersections. Compared to baseline regression models such as KNN, SVR, MLP, AdaBoost, XGBoost, CatBoost, and LightGBM, the proposed framework consistently outperformed them, achieving significant improvements in estimation accuracy. For example, the MAE for the through movement using the proposed framework (34.39) is substantially lower than that of SVR (93.71), KNN (84.35), and MLP (60.12). Similarly, for left-turn movements, the proposed model's MAE (12.29) is better than CatBoost (13.72) and LightGBM (13.24). For right-turn movements, the proposed framework also outperformed others, with an MAE of 16.17 compared to CatBoost (18.04) and LightGBM (18.59).

In addition to the baseline regression models, the table also includes five domain adaptation models: ITML-IWC, ITML-IWN, ITML-RULSIF, ITML-WANN, and ITML-GBBW. Among these, ITML-GBBW demonstrated the best performance, with MAE values of 12.43 for left-turn, 35.83 for through, and 16.74 for right-turn movements. While ITML-GBBW performed well, the proposed ITMLGMM-GBBW model further improved estimation accuracy, reducing the MAE by 1.13% for left-turn, 4.02% for through, and 3.41% for right-turn movements compared to ITML-GBBW. Other domain adaptation models, such as ITML-IWC, ITML-IWN, ITML-RULSIF, and ITML-WANN, showed competitive performance but were less accurate than the proposed model.

**Table 3 MAE Comparison for estimating TMCs of different movements**

| Model | Left-turn | Through | Right-turn |
|---|---|---|---|
| KNN | 18.85 | 84.35 | 21.86 |
| SVR | 17.28 | 93.71 | 19.93 |
| MLP | 18.19 | 60.12 | 20.47 |
| AdaBoost | 13.96 | 43.53 | 18.60 |
| XGBoost | 13.79 | 37.97 | 19.78 |
| CatBoost | 13.72 | 38.36 | 18.04 |
| LightGBM | 13.24 | 36.48 | 18.59 |
| ITML-IWC | 13.91 | 42.13 | 17.29 |
| ITML-IWN | 14.18 | 42.04 | 17.50 |
| ITML-RULSIF | 13.92 | 42.15 | 17.30 |
| ITML-WANN | 14.36 | 36.36 | 17.62 |
| ITML-GBBW | 12.43 | 35.83 | 16.74 |
| **ITMLGMM-GBBW** | **12.29** | **34.39** | **16.17** |

Table 4 provides a comprehensive comparison of estimation performance, measured in terms of Root Mean Square Error (RMSE), across seven state-of-the-art regression models, five domain adaptation models, and the proposed ITMLGMM-GBBW model, for estimating TMCs across three movement types: left-turn, through, and right-turn. The RMSE values represent the average performance across 30 intersections for each movement type.

The proposed ITMLGMM-GBBW model, achieved the lowest RMSE values among all models, with 15.73 for left-turn, 43.90 for through, and 20.08 for right-turn movements. These results highlight its superior performance in estimating TMCs at intersections. When compared to state-of-the-art regression models such as KNN, SVR, MLP, AdaBoost, XGBoost, CatBoost, and LightGBM, the proposed framework consistently outperformed them, achieving significant reductions in RMSE. For instance, the RMSE for the through movement using the proposed framework (43.90) is dramatically lower than that of SVR (107.69), KNN (97.18), and MLP (72.01). Similarly, for left-turn movements, the proposed model's RMSE (15.73) is better than CatBoost (17.41) and LightGBM (16.97). For right-turn movements, the proposed framework also demonstrated superior accuracy, with an RMSE of 20.08 compared to LightGBM (22.50) and CatBoost (21.79).

In addition to outperforming state-of-the-art regression models, the proposed framework also surpassed five domain adaptation models: ITML-IWC, ITML-IWN, ITML-RULSIF, ITML-WANN, and ITML-GBBW. Among these, ITML-GBBW showed the best performance, with RMSE values of 16.03 for left-turn, 45.21 for through, and 20.77 for right-turn movements. However, the proposed ITMLGMM-GBBW model further improved estimation accuracy, reducing the RMSE by 1.87% for left-turn, 2.90% for through, and 3.32% for right-turn movements compared to ITML-GBBW. Other domain adaptation models, such as ITML-IWC, ITML-IWN, ITML-RULSIF, and ITML-WANN, demonstrated competitive performance but were less accurate than the proposed model.

**Table 4 RMSE Comparison for estimating TMCs of different movements**

| Model | Left-turn | Through | Right-turn |
|---|---|---|---|
| KNN | 23.87 | 97.18 | 26.83 |
| SVR | 22.07 | 107.69 | 24.17 |
| MLP | 22.58 | 72.01 | 24.60 |
| AdaBoost | 17.77 | 54.06 | 22.88 |
| XGBoost | 18.03 | 48.75 | 24.09 |
| CatBoost | 17.41 | 47.63 | 21.79 |
| LightGBM | 16.97 | 46.40 | 22.50 |
| ITML-IWC | 17.48 | 51.28 | 20.96 |
| ITML-IWN | 17.86 | 51.47 | 21.02 |
| ITML-RULSIF | 17.49 | 51.32 | 20.98 |
| ITML-WANN | 17.78 | 45.43 | 21.64 |
| ITML-GBBW | 16.03 | 45.21 | 20.77 |
| **ITMLGMM-GBBW** | **15.73** | **43.90** | **20.08** |

Figure 4 compares the Mean Absolute Error (MAE) of various models for estimating TMCs across three movement types: left-turn, through, and right-turn. Across all movement types, ITMLGMM-GBBW consistently achieved the lowest MAE with minimal variability, demonstrating its superior accuracy and robustness.

For left-turn movements, ITMLGMM-GBBW outperformed all other models, achieving the lowest MAE with a tightly clustered distribution, indicating consistent performance. State-of-the-art regression models such as KNN, SVR, and MLP exhibited higher MAE values and greater variability, while domain adaptation models like ITML-GBBW and ITML-WANN performed better but were still less accurate than the proposed model.

Through movements, the most challenging estimation task, showed higher MAE values and greater variability across all models. Despite this, ITMLGMM-GBBW achieved the lowest median MAE and a relatively narrow distribution, outperforming both state-of-the-art regression models and domain adaptation methods. While domain adaptation models like ITML-GBBW and ITML-WANN performed competitively, they were consistently outperformed by the proposed model. State-of-the-art regression models such as SVR and KNN struggled the most, with significantly higher MAE values and wider variability.

For right-turn movements, ITMLGMM-GBBW again demonstrated superior performance, achieving the lowest MAE with a tightly clustered distribution. State-of-the-art regression models showed higher MAE and greater variability, while domain adaptation models like ITML-GBBW and ITML-RULSIF performed better but still fell short of the proposed model's accuracy.

Overall, ITMLGMM-GBBW consistently outperformed all other models across all movement types, demonstrating its ability to deliver accurate and reliable TMC estimates. State-of-the-art regression models struggled with higher MAE and variability, particularly for through movements, while domain adaptation models showed improved performance but were still less effective than the proposed approach. These results highlight the effectiveness of ITMLGMM-GBBW in addressing the complexities of traffic estimation, showcasing the potential of ITMLGMM-GBBW to enhance accuracy and consistency in real-world applications.

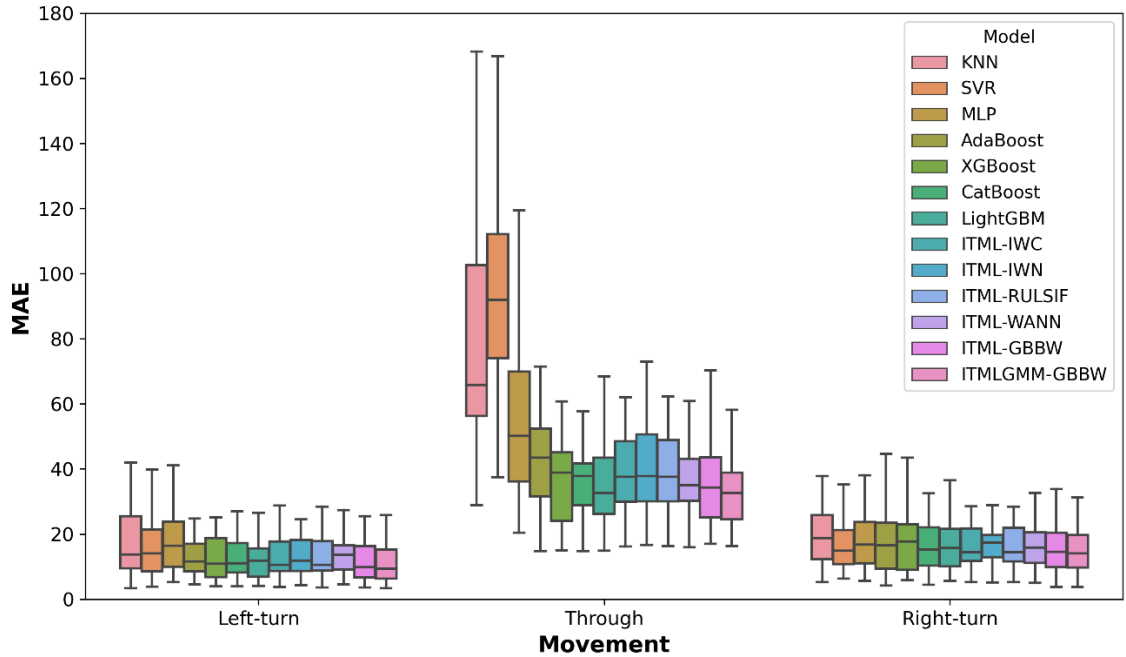

**Figure 4 MAE comparison between different models for different movements**

Figure 5 compares the Root Mean Square Error (RMSE) of various models for estimating TMCs across left-turn, through, and right-turn movements. The proposed DA framework consistently achieved the lowest RMSE with minimal variability, outperforming both state-of-the-art regression models (e.g., KNN, SVR, MLP) and domain adaptation models (e.g., ITML-GBBW, ITML-WANN).

Similar to Figure 4, for left-turn and right-turn movements, ITMLGMM-GBBW demonstrated superior accuracy with tightly clustered RMSE distributions. In contrast, state-of-the-art regression models showed higher RMSE and greater variability. For through movements, the most challenging case, ITMLGMM-GBBW again achieved the lowest RMSE, with domain adaptation models performing better than state-of-the-art regression models but still falling short of the proposed model.

Overall, ITMLGMM-GBBW consistently outperformed all other models, demonstrating its robustness and accuracy in estimating TMCs across all movement types. This highlights the effectiveness of DA in addressing complex traffic estimation challenges.

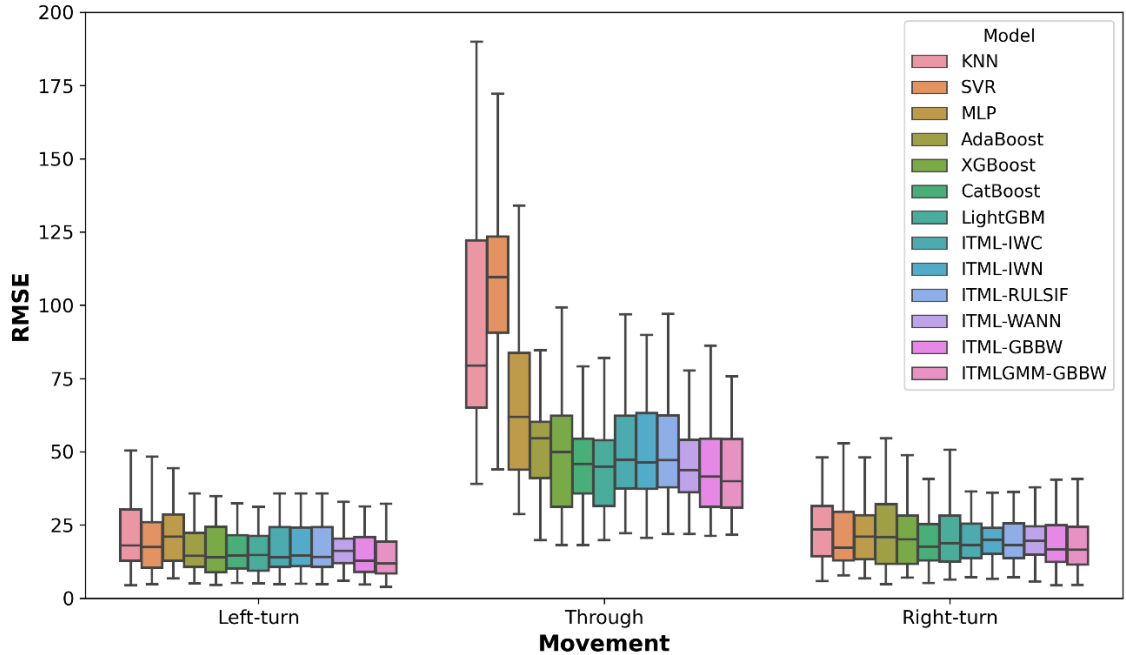

**Figure 5 RMSE comparison between different models for different movements**

Compared to state-of-the-art regression models, the proposed framework achieves the highest accuracy in estimating TMCs by relaxing the assumption that the source and target domains must share identical data distributions. ITMLGMM-GBBW effectively handles variations in traffic patterns, distributions, and characteristics, enabling the development of scene-specific models that outperform generic approaches.

These findings further emphasize the superiority of the proposed ITMLGMM-GBBW model, which consistently outperforms not only state-of-the-art regression models but also advanced domain adaptation methods. By incorporating a GMM-based data augmentation strategy, the framework enhances generality and adaptability, enabling it to deliver substantial reductions in both MAE and RMSE across all movement types. This highlights its ability to provide more precise and reliable TMC estimations at intersections. The integration of GMM demonstrates a clear advantage over other advanced domain adaptation methods in addressing the complexities of traffic estimation in urban environments.

### 4.5. Ablation Study

#### 4.5.1. Ablation Study on Components and Generated Samples of GMM

Figure 6 provides a detailed evaluation of the performance of the Gaussian Mixture Model (GMM) in generating synthetic samples for estimating movement counts across three traffic movement types: left-turn, through, and right-turn. The performance is assessed using two error metrics: Mean Absolute Error (MAE) and Root Mean Squared Error (RMSE), which are calculated separately for each movement type. These metrics provide insight into the accuracy and reliability of the synthetic data generated by the GMM.

The analysis investigates how the number of components in the GMM (n_components) and the number of generated samples (n_samples) influence the average MAE and RMSE. The parameter n_components determines the number of Gaussian distributions (or components) used by the GMM to model the data, with each component representing a distinct Gaussian distribution. The GMM combines these components as a weighted sum to approximate the underlying data distribution. The x-axis of the plot represents the number of components in the GMM, ranging from 1 to 5, which controls the model's complexity and its ability to capture the data's structure.

The y-axis displays the average MAE and RMSE values, where lower values indicate better performance in terms of accurately estimating movement counts. The legend provides additional context by showing the number of samples generated by the GMM for each configuration, ranging from 20 to 200 in increments of 20. This allows for a comprehensive analysis of how both the model complexity (n_components) and the volume of synthetic data (n_samples) impact the error metrics.

The results reveal that the MAE varies across different values of n_components and n_samples. Based on the observation, n_components = 2 and n_samples = 40 is set for left-turn movement TMC estimation, n_components = 4 and n_samples = 100 is set for through movement TMC estimation, and n_components = 5 and n_samples = 180 is set for right-turn movement TMC estimation. These findings provide valuable guidance for selecting GMM parameters to achieve accurate traffic movement estimation.

By including separate evaluations for left-turn, through, and right-turn movements, the plot offers a granular view of the GMM's performance across different traffic movement types, enabling a more nuanced understanding of its effectiveness in generating realistic synthetic data.

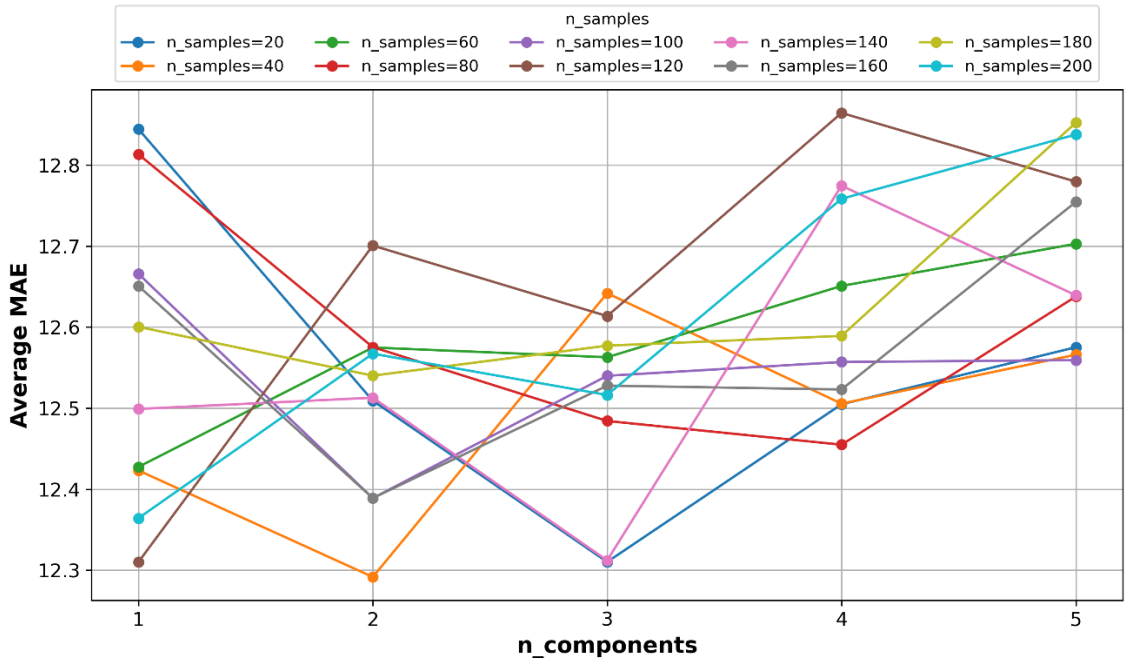

(a) Left-turn movement average MAEs across different components and generated samples

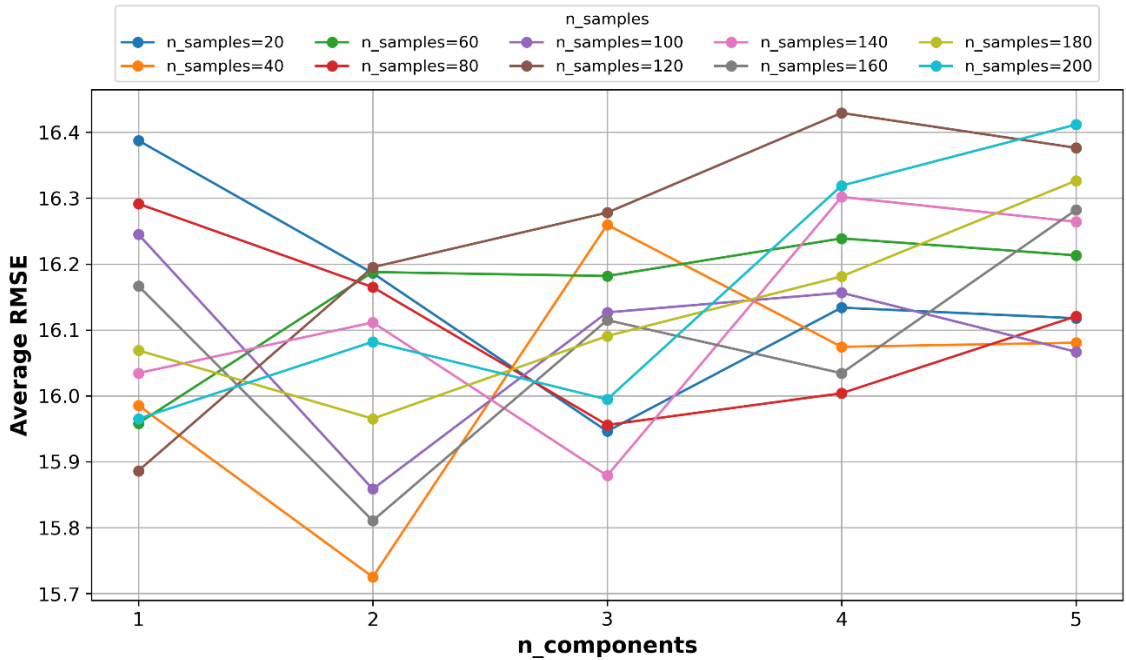

(b) Left-turn movement average RMSEs across different components and generated samples

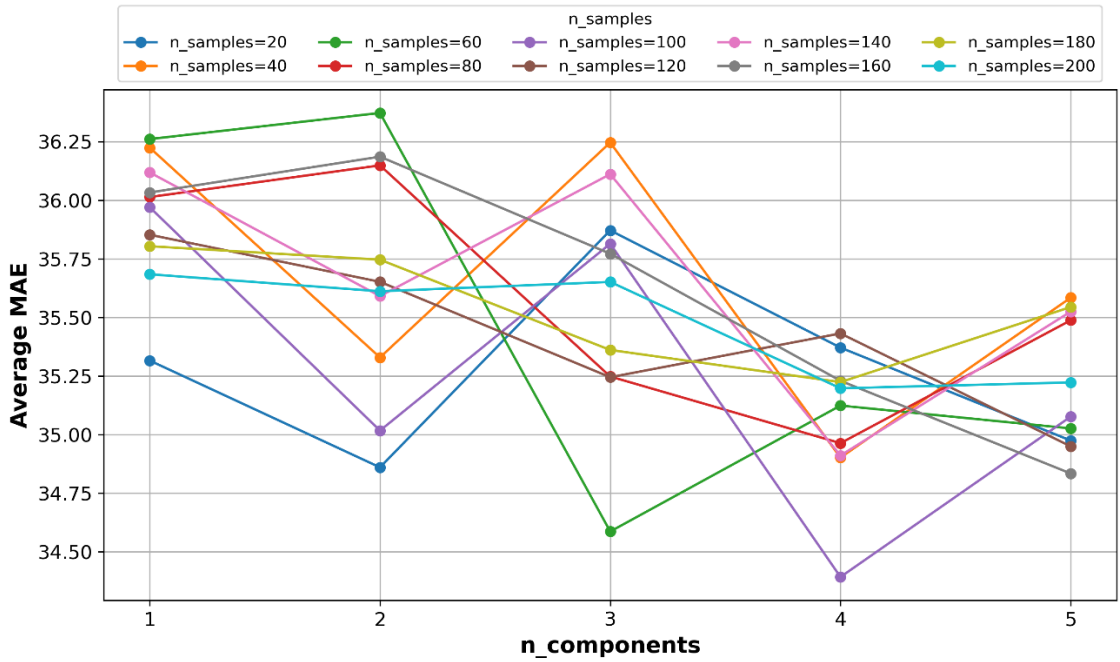

(c) Through movement average MAEs across different components and generated samples

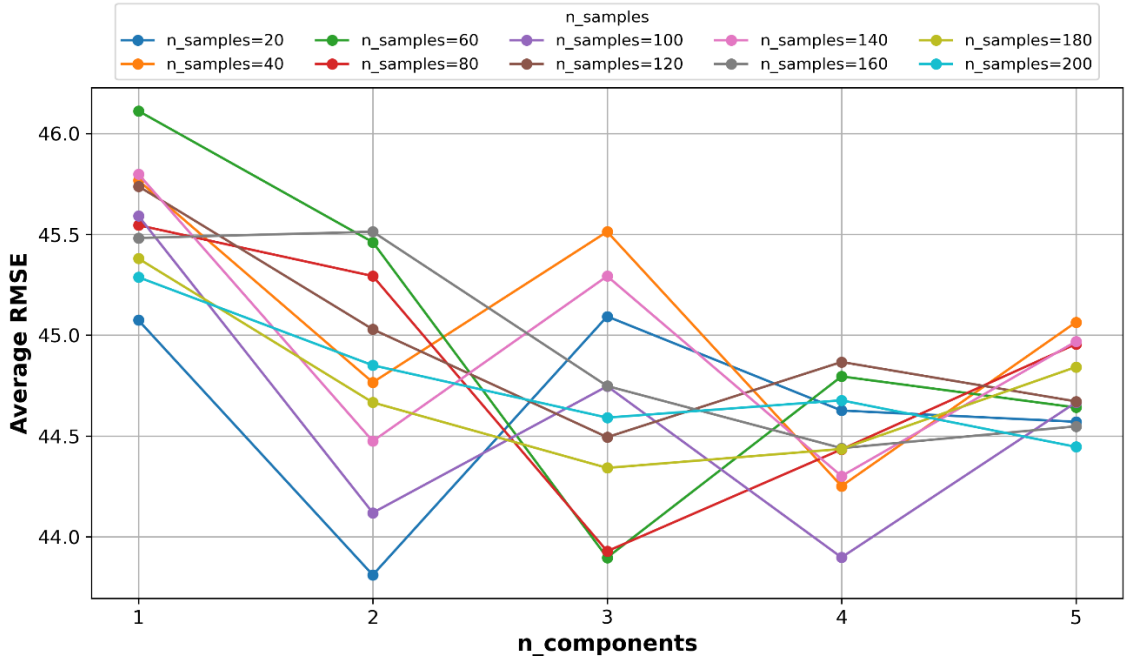

(d) Through movement average RMSEs across different components and generated samples

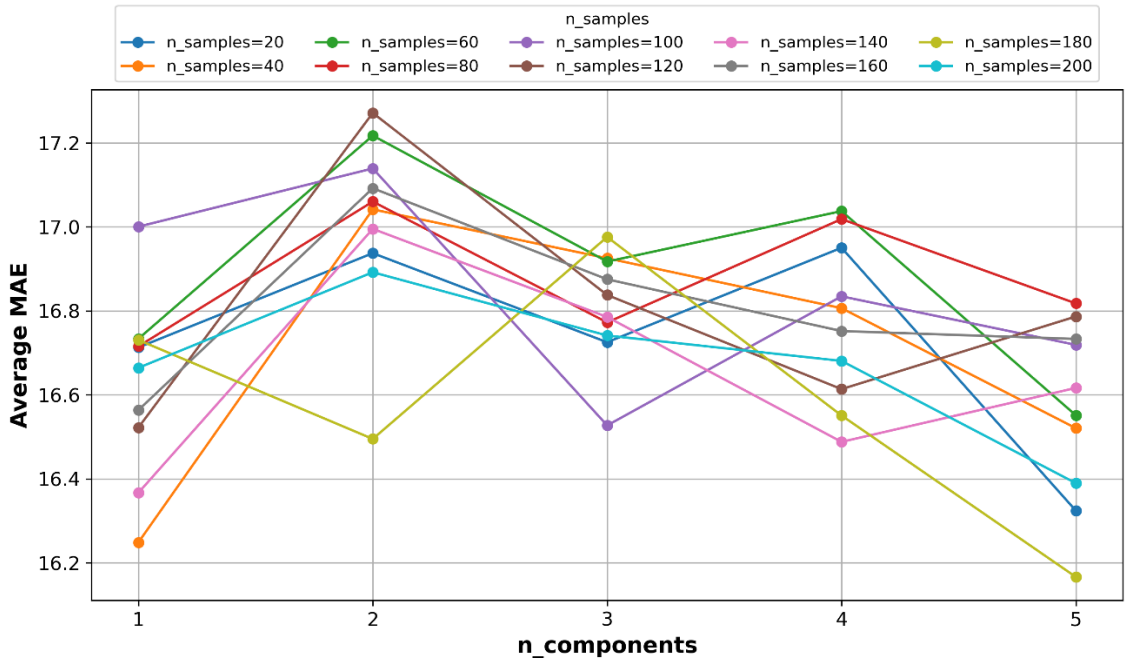

(e) Right-turn movement average MAEs across different components and generated samples

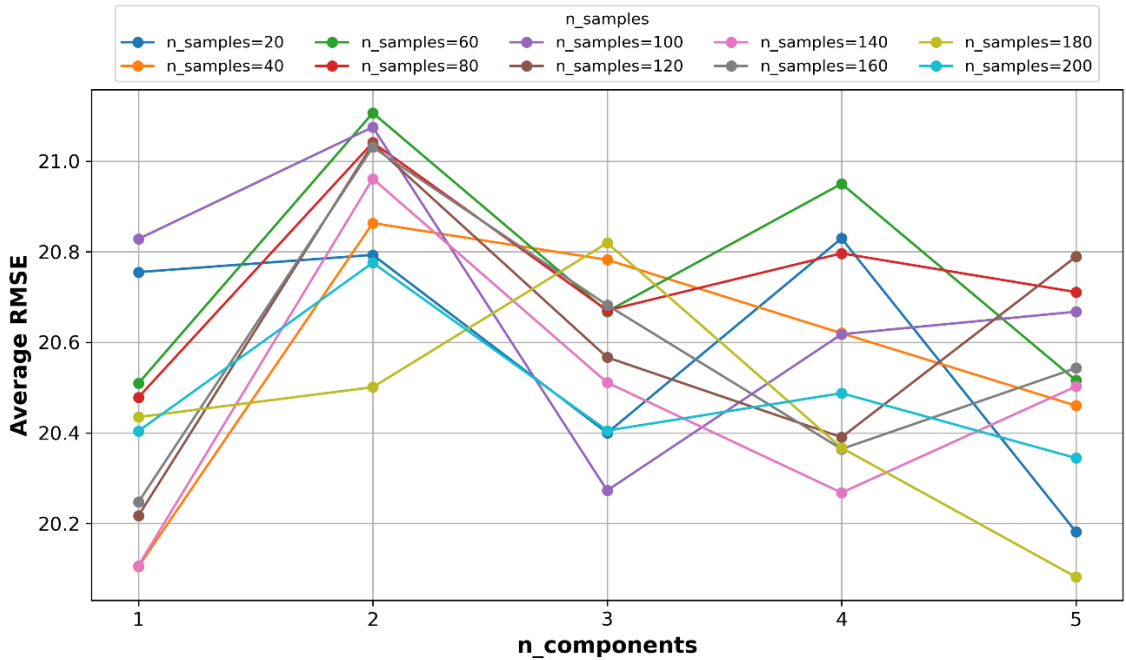

(f) Right-turn movement average RMSEs across different components and generated samples

**Figure 6 Average MAEs and RMSEs across different components and generated samples**

### 4.5.2. Ablation Study on Alpha Values

Figure 7 demonstrates the relationship between the alpha parameter (denoted as $\alpha$ in the description) and the average MAEs and RMSEs in the context of using the proposed framework to estimate TMCs. The x-axis represents the alpha parameter, which ranges from 0 to 1.0, while the y-axis shows the average MAEs and RMSEs, Lower MAE and RMSE values indicate better model performance.

The alpha parameter determines the balance between the contributions of the source and target domains during the learning process. When alpha equals 0, the model relies entirely on the source domain data, effectively ignoring the target domain. In this case, the source data is weighted fully ($1 - \alpha = 1$), while the target data is not considered ($\alpha = 0$). This approach may work well if the source and target domains are very similar, but it risks poor generalization if there is a significant domain shift, as the model does not incorporate any information from the target domain. When alpha equals 0.5, the model gives equal importance to both the source and target domains, with both being weighted equally ($1 - \alpha = 0.5$ and $\alpha = 0.5$). This balanced approach is ideal when the source and target domains are moderately similar but still have some differences, as it allows the model to leverage general patterns from the source domain while adapting to domain-specific characteristics in the target domain. Finally, when alpha equals 1, the model relies entirely on the target domain data, ignoring the source domain. In this case, the target data is fully weighted ($\alpha = 1$), while the source data is not considered ($1 - \alpha = 0$). This approach may be beneficial if the target domain is sufficiently large and representative of the task, but it discards potentially useful information from the source domain, which could improve generalization, especially if the target data is limited or noisy.

For alpha values between 0 and 0.9, the average MAEs and RMSEs remain relatively stable. This suggests that the model performs consistently well in this range. However, as alpha approaches 1.0, the MAEs and RMSEs spike dramatically. This sharp increase indicates a significant decline in model performance, likely due to overemphasis on target data, which disrupts the balance and leads to poor generalization. The plot suggests that alpha plays a critical role in balancing the source and target domain instances or weighting certain features in the DA process. Lower alpha values (0–0.9) maintain a good balance, resulting in stable and low MAEs and RMSEs, while higher alpha values (1.0) disrupt this balance, causing performance degradation. Overall, a balanced approach ($\alpha = 0.5$) is often preferred for domain adaptation tasks, as it allows the model to effectively leverage information from both domains.

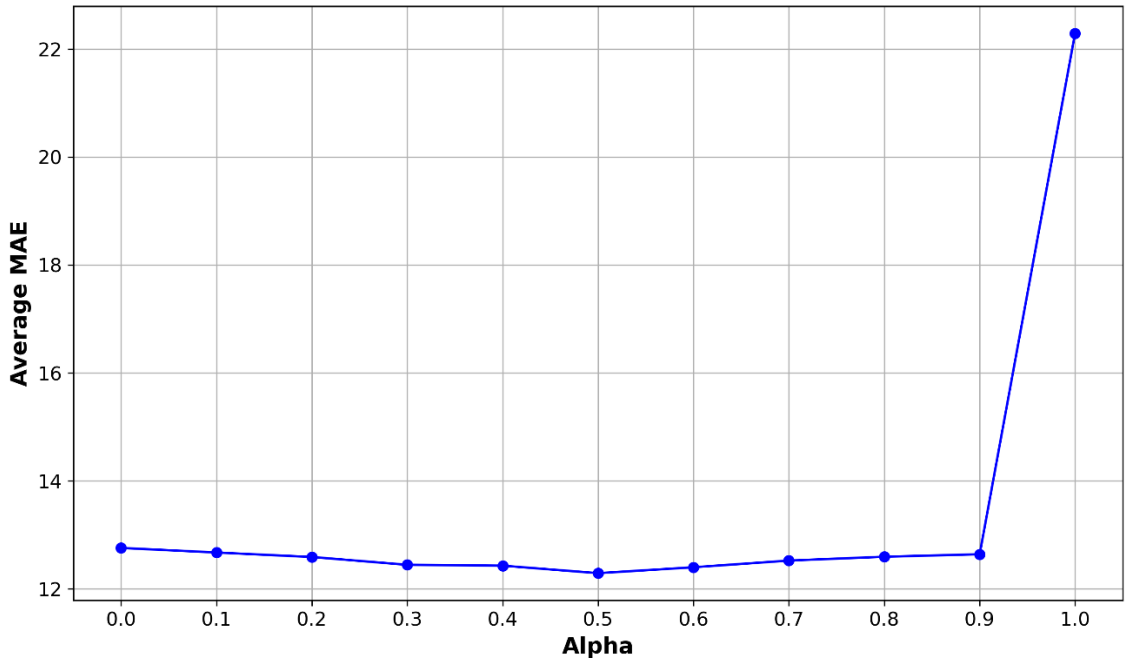

(a) Left-turn movement average MAEs across different alpha values

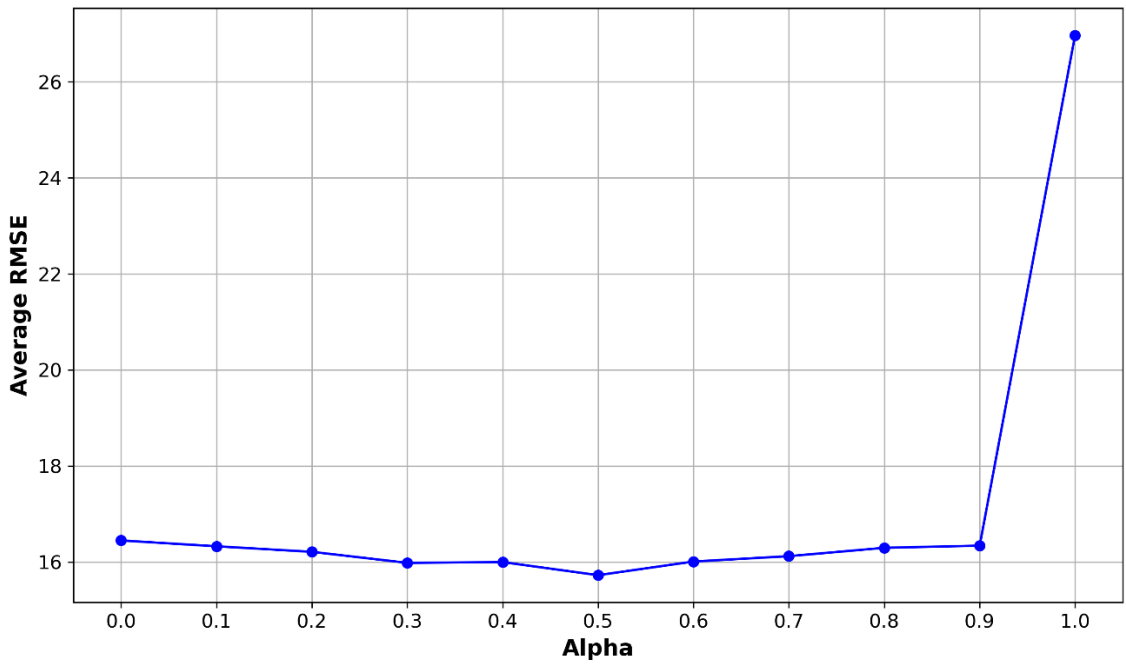

(b) Left-turn movement average RMSEs across different alpha values

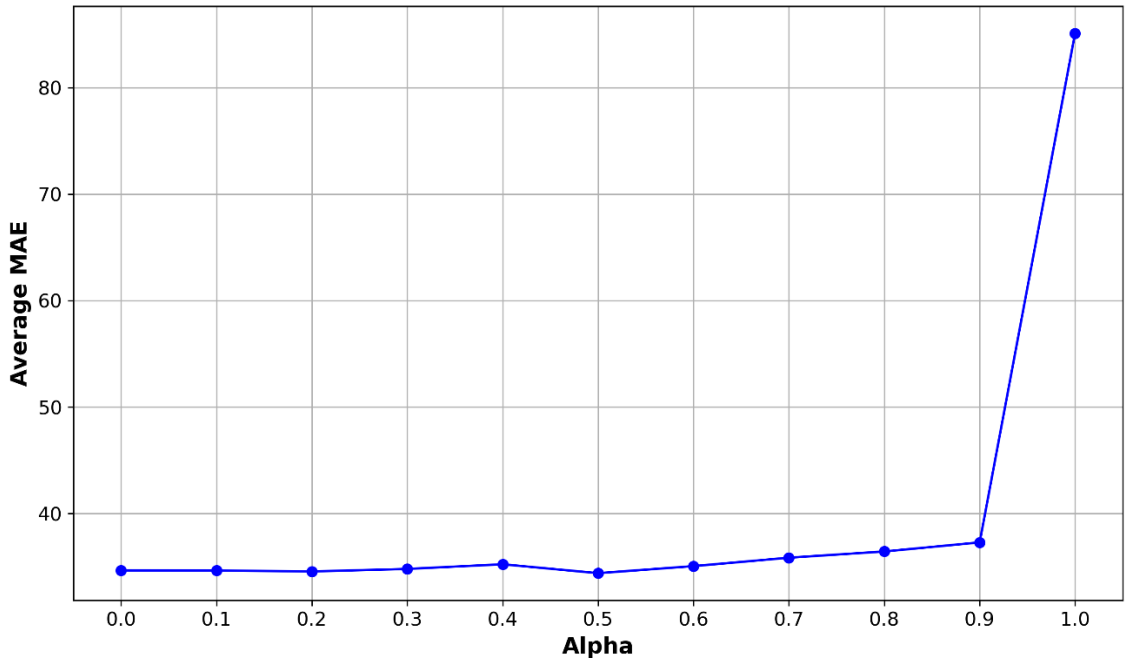

(c) Through movement average MAEs across different alpha values

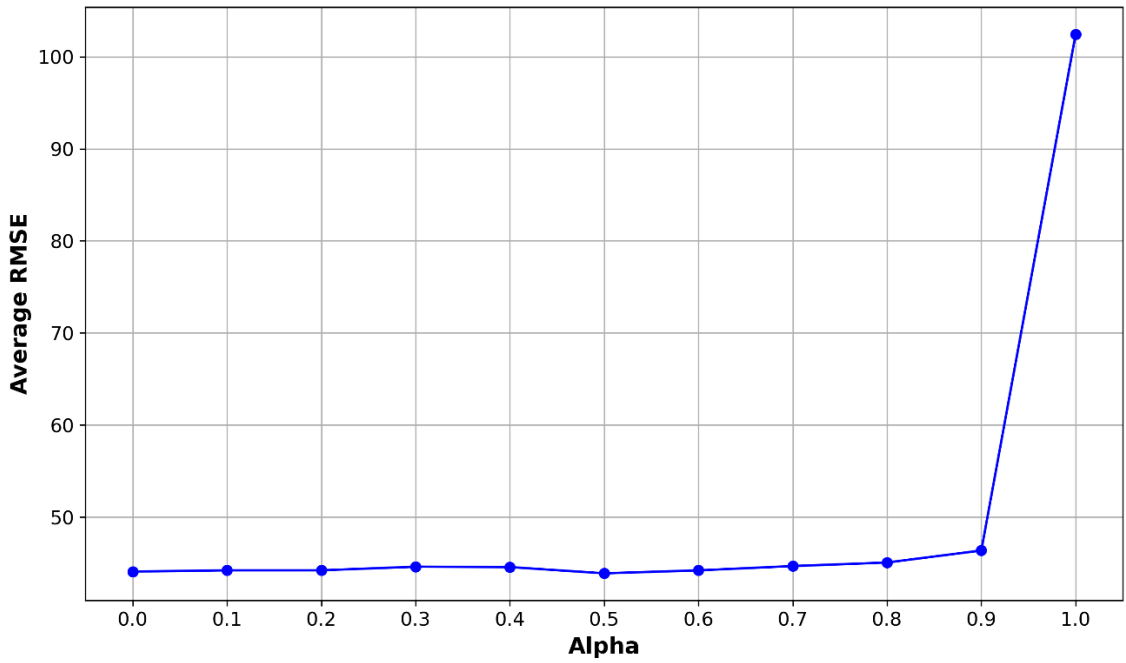

(d) Through movement average RMSEs across different alpha values

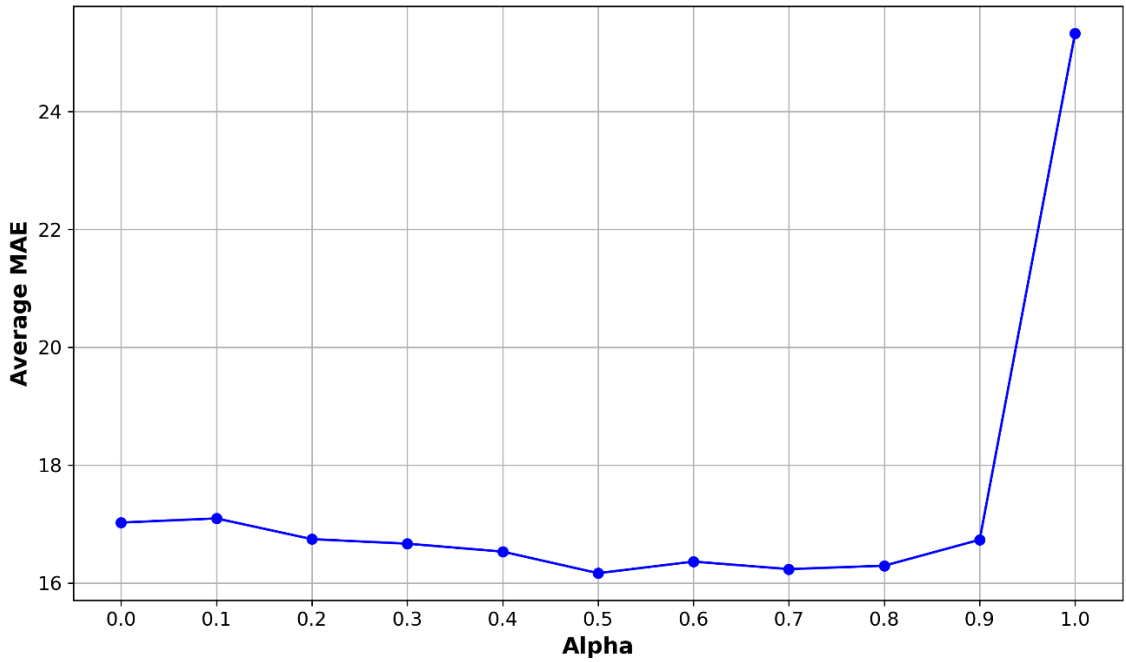

(e) Right-turn movement average MAEs across different alpha values

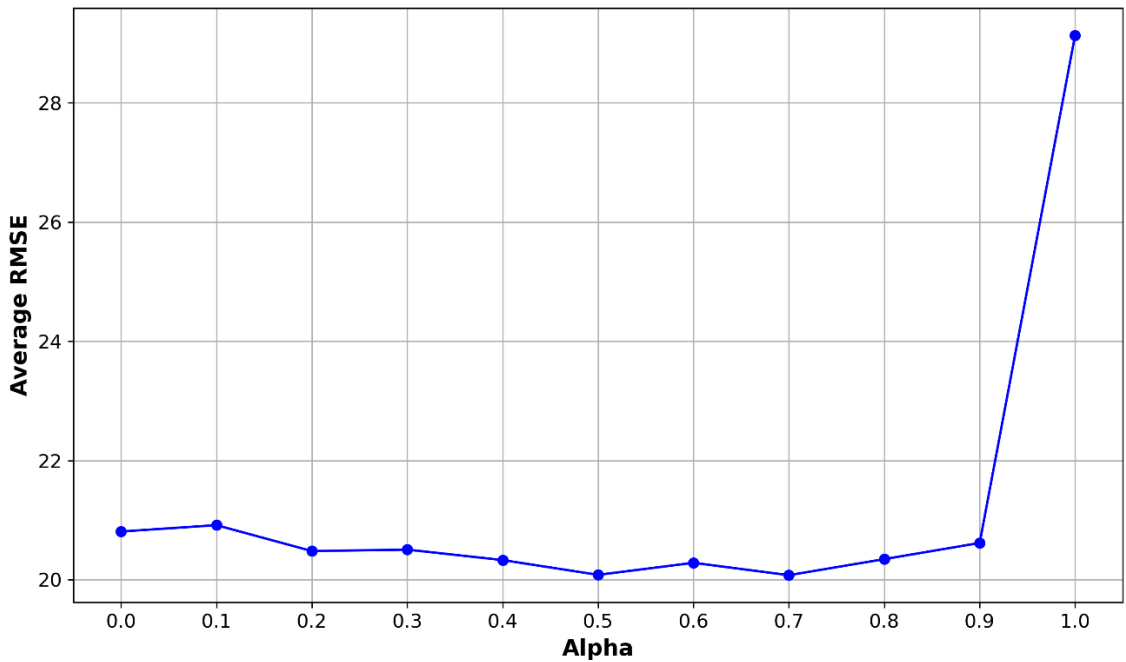

(f) Right-turn movement average RMSEs across different alpha values

**Figure 7 Average MAEs and RMSEs across different alpha values**

# 5. CONCLUSION

This research proposes a DA framework for TMC estimation, leveraging data from well-instrumented intersections to infer traffic patterns at intersections lacking sensors. Using traffic controller event-based data, road infrastructure data, and point-of-interest data, the DA framework adapts to diverse traffic patterns, ensuring high accuracy and scalability. Evaluated on 30 intersections in Tucson, Arizona, the proposed framework stood out by achieving the lowest MAE values of 9.57 for left-turn, 28.59 for through, and 10.25 for right-turn movements and the lowest RMSE values of 13.65 for left-turn, 40.62 for through, and 13.98 for right-turn movements, indicating its effectiveness in estimating TMCs at intersections.

This study introduces a pioneering DA framework for TMC estimation, demonstrating its potential to enhance accuracy and efficiency in traffic management. By rigorously comparing the proposed framework to traditional models, the research highlights its advantages while addressing potential challenges. The framework provides precise, scene-specific TMC estimates tailored to individual intersections, enabling more effective traffic management decisions. Additionally, it offers a scalable and cost-effective solution by leveraging traffic controller event data, which ensures region-wide coverage and adaptability to dynamic urban traffic conditions. By promoting data efficiency and resource optimization, the DA framework reduces the dependency on extensive datasets and costly physical sensors, making it a practical and sustainable approach for modern traffic systems.

As the framework has high flexibility allowing for the incorporation of multiple variables, future work could consider adding more temporal and spatial variables to the proposed model. Also, the weather conditions, events or incidents, and other factors that might impact the traffic operation can be included in the framework to further improve the generalization ability of the model. This paper is the first attempt to employ DA for TMC estimation. In the future, more advanced machine learning methods and more sophisticated input features could be applied to further improve the estimation performance.